\pdfoutput=1

\documentclass[11pt]{article}

\usepackage[final]{acl}

\usepackage{times}
\usepackage{latexsym}
\usepackage{subcaption}
\usepackage[most]{tcolorbox}

\usepackage[T1]{fontenc}

\usepackage[utf8]{inputenc}

\usepackage{microtype}

\usepackage{inconsolata}

\usepackage{graphicx}

\usepackage{amsmath}
\usepackage{amssymb}
\usepackage{mathtools}
\usepackage{amsthm}

\usepackage{microtype}
\usepackage{graphicx}
\usepackage{booktabs} 
\usepackage{enumitem}
\usepackage{siunitx}
\usepackage{listings}
\usepackage{xcolor}
\usepackage{multirow}
\usepackage{graphicx}
\usepackage{float}
\usepackage[textsize=tiny]{todonotes}

\title{Assay2Mol: Large Language Model-based Drug Design Using BioAssay Context}

\author{\textbf{Yifan Deng\textsuperscript{1, 2}},  \textbf{Spencer S. Ericksen\textsuperscript{3}, \textbf{Anthony Gitter\textsuperscript{1,2,4}}} \\
 \textsuperscript{1} Department of Computer Sciences, University of Wisconsin-Madison\\
 \textsuperscript{2} Morgridge Institute for Research\\
 \textsuperscript{3} Drug Development Core, Small Molecule Screening Facility, \\University of Wisconsin Carbone Cancer Center, University of Wisconsin-Madison\\
 \textsuperscript{4} Department of Biostatistics and Medical Informatics, University of Wisconsin-Madison\\
 \texttt{\{yifan.deng,ssericksen\}@wisc.edu}\\
 \texttt{gitter@biostat.wisc.edu}
}

\begin{document}
\maketitle
\begin{abstract}
Scientific databases aggregate vast amounts of quantitative data alongside descriptive text. In biochemistry, molecule screening assays evaluate candidate molecules' functional responses against disease targets. Unstructured text that describes the biological mechanisms through which these targets operate, experimental screening protocols, and other attributes of assays offer rich information for drug discovery campaigns but has been untapped because of that unstructured format. We present Assay2Mol, a large language model-based workflow that can capitalize on the vast existing biochemical screening assays for early-stage drug discovery. Assay2Mol retrieves existing assay records involving targets similar to the new target and generates candidate molecules using in-context learning with the retrieved assay screening data. Assay2Mol outperforms recent machine learning approaches that generate candidate ligand molecules for target protein structures, while also promoting more synthesizable molecule generation.
\end{abstract}

\section{Introduction}
Early-stage drug development and target validation typically involve a search through chemical space for drug-like molecules that perturb a target of interest, usually a protein activity connected to a disease condition. For a new target, the search starts with assay development and scaling for high-throughput testing in order to experimentally screen a pre-defined chemical library for active molecules. From data obtained on screens of a target of interest (or related targets), computational models can learn structure-activity relationships that inform selection of new molecules for testing.

Public screening data repositories like PubChem \citep{10.1093/nar/gkae1059} and ChEMBL \citep{10.1093/nar/gky1075} possess great value in this regard. PubChem now contains 1.77 million assay records (BioAssay records) comprising $\sim$300 million bioactivity outcomes across $\sim$250,000 protein targets and $\sim$2,000 cell lines\footnote{\url{https://pubchem.ncbi.nlm.nih.gov/docs/statistics}}. BioAssay records are richly annotated with chemicals tested, target genes, pathways, proteins, cell lines, publications, patents, related BioAssay records, tabular molecule testing results, text descriptions of assay format, protocols, and relevance to disease states.

Sorting through these repositories for the data pertinent to an arbitrary target is a daunting task. Scientists need workflows that can rapidly identify relevant BioAssay records based on their associated text, extract key textual and tabular chemical testing data comprising molecule structures paired with experimental activity outcomes, and apply this information to models capable of learning structure-activity relationships to recommend new molecules for testing.

Given the extensive descriptive text components in BioAssay records, leveraging natural language processing capabilities becomes crucial for efficient retrieval and interpretation. Large language models (LLMs), with their advanced ability to process and interpret unstructured text, are well-suited for assessing BioAssay relevance by extracting key experimental details and identifying meaningful activity patterns. LLMs have demonstrated great ability in different kinds of tasks including translation, multi-round conversation, and so on. LLMs are also adept in biology- and chemistry-related tasks, for example, molecule property prediction, and text-guided molecule generation. 
LLMs support in-context learning, which extends to scientific domains. GPT-3 \citep{brown_language_2020} highlights the ability of LLMs to adapt to new tasks with minimal examples. In text-guided molecule design, GPT-4 outperforms other fine-tuned models with few-shot in-context learning \citep{zhao_chemdfm_2024, guo_what_2023}. This raises the question of whether LLMs can support different strategies for molecule design. Beyond generating molecules that satisfy specified property constraints, can LLMs navigate unstructured text within public BioAssay records and then use it as context to design molecules with functional properties, such as protein inhibition?

\begin{figure*}[!ht]
    \centering
    \includegraphics[width=0.9\linewidth]{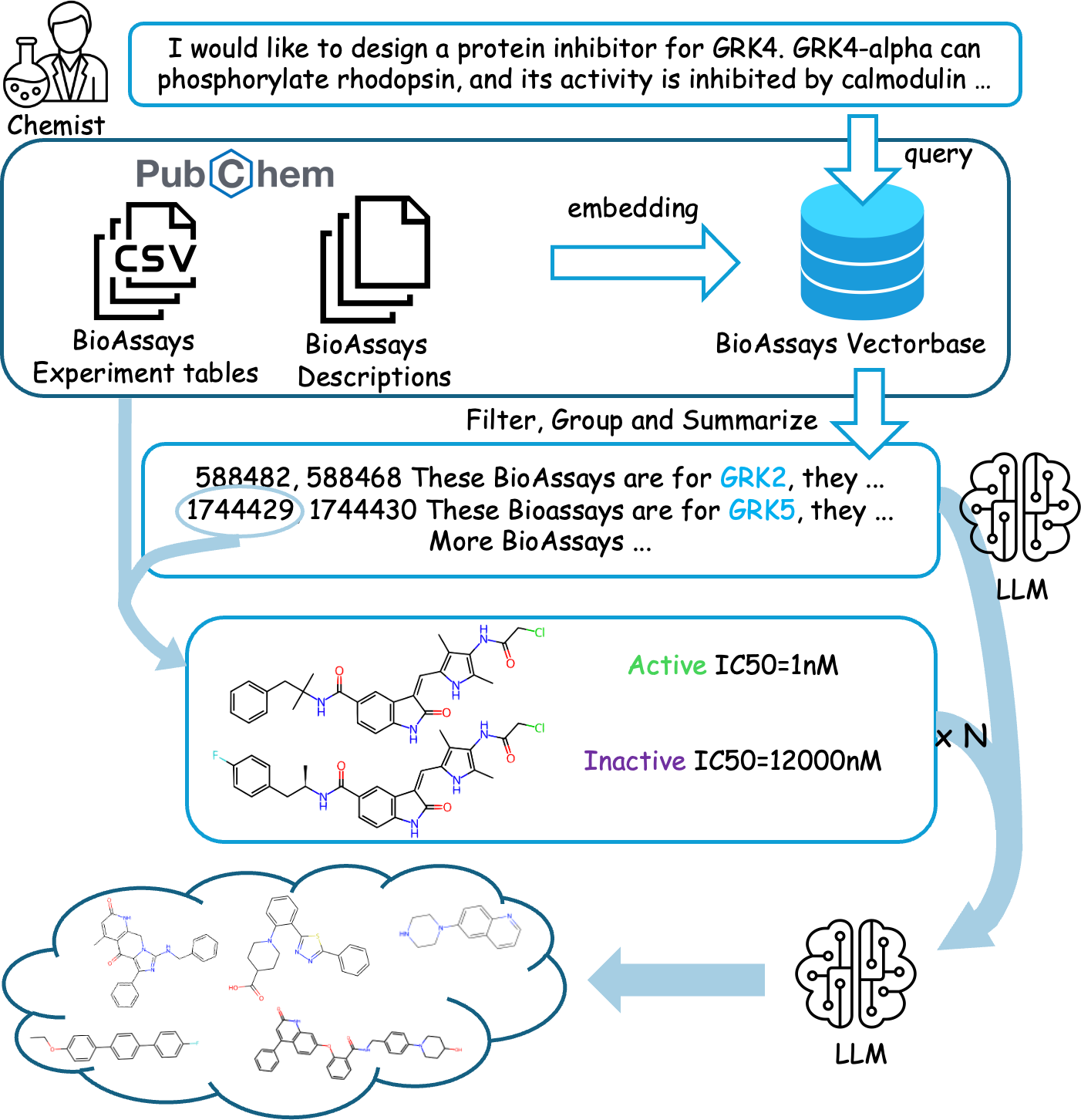}
    \caption{The Assay2Mol workflow. A chemist provides a target description, which is used to retrieve BioAssays from the pre-embedded vector database. After filtering for relevance, the BioAssays are summarized by an LLM. The BioAssay ID is then used to retrieve experimental tables. The final molecule generation prompt is formed by combining the description, summarization, and selected test molecules with associated test outcomes, enabling the LLM to generate relevant active molecules. Icons are from Flaticon.com and svgrepo.com}
    \label{fig:workflow}
\end{figure*}

We propose Assay2Mol to maximize the use of BioAssay records with LLMs. Given input such as protein target descriptions, phenotypic data, or other textual information, Assay2Mol retrieves relevant BioAssays and leverages in-context learning to generate molecules with the desired biological activity (Figure \ref{fig:workflow}). Our contributions can be summarized as follows:
\begin{itemize}[noitemsep, topsep=0pt]
    \item We introduce Assay2Mol, an LLM-based drug design workflow that retrieves relevant PubChem BioAssay data for a given query and then learns to generate candidate molecules from this assay context.
    \item Unlike structure-based drug discovery (SBDD) methods, Assay2Mol does not require protein structures or even sequences. It can even generate candidate active molecules for cell-based and phenotypic assays and endpoints (e.g., tumor shrinkage, cardiotoxicity, QT interval prolongation).
    \item Because Assay2Mol relies on LLMs that include molecules in their pretraining data, some output molecules are more like "retrieval" rather than \textit{de novo} "generation". This increases the chemical plausibility and synthetic accessibility of generated molecules.
\end{itemize}

\section{Related work}
\subsection{Large Language Models in biochemistry}
LLMs are an alternative to biomolecular sequence- or structure-based models for learning structure-activity or structure-property relationships. Multimodal molecule and text generation \citep{edwards_text2mol_2021, pei-etal-2024-biot5, deng_chemical_2024, zhao_chemdfm_2024, fang2023mol}, in-context learning for chemistry \citep{fifty_context_2023, jablonka_leveraging_2024, nguyen_lico_2024, moayedpour_many-shot_2024, schimunek_mhnfs_2025}, and LLM agents \citep{M.Bran2024, doi:10.1021/acsomega.4c08408, swanson_virtual_2024, gao_pharmagents_2025, liu_drugagent_2025, wei_fleming_2025} are at the frontier of this interface of LLMs and biomolecules. Recent reviews \cite{zhang_scientific_2024, mirza_are_2024, ramos_review_2025, wang_survey_2025, alampara_general_2025} provide broader coverage of this expansive area.

\subsection{BioAssay data mining}
BioAssay databases like PubChem and ChEMBL are valuable resources for data mining and have been used to train many machine learning models. \citet{sharma_data_2024} developed a data mining pipeline that compiled and processed 8,415 OXPHOS-related BioAssays from PubChem, identified major OXPHOS inhibitory chemotypes, and trained effective OXPHOS inhibitor classifiers. MolecularGPT \citep{liu_moleculargpt_2024} constructed an instruction tuning dataset by collecting three-shot examples from ChEMBL. MBP \cite{10.1093/bib/bbad451} created a multi-task pretraining dataset with labels from BioAssays to address label inconsistencies and data scarcity. 
\citet{10.5555/3618408.3619671} proposed CLAMP, a dual‐encoder architecture combining chemical structure and natural‐language descriptions of BioAssays, trained with a contrastive objective to enable zero‐ and few‐shot activity prediction.
TwinBooster \cite{doi:10.1021/acs.jcim.4c00765} is a zero-shot model that integrates molecule structures and BioAssay descriptions for molecular property prediction. 
\citet{schoenmaker_toward_2025} demonstrated that incorporating assay-aware embeddings derived from ChEMBL assay descriptions can reduce variance in bioactivity data and improve proteochemometric modeling performance. \citet{doi:10.1021/acs.jcim.4c02122} extracted structured BioAssay data from ChEMBL, PubChem, and literature to train AutoML-based models for 11 ADMET properties, without employing any natural language information.
\citet{smit_enhancing_2025} developed manual and AI-based methods to improve BioAssay annotations in ChEMBL, including automated extraction of experimental methods and refined classification. It increased the reusability of bioactivity data, supporting more reliable downstream modeling.

\subsection{Structure-based ligand design}
Molecule design approaches play a role in the hit finding and lead optimization tasks of early-stage drug discovery. The goal of hit finding is to identify pharmacologically active molecules for a target of interest to serve as starting points for development. Given availability of a 3D structure model for a protein target, structure-based virtual screening methods can be applied for hit finding, such as large-scale docking of pre-enumerated molecule libraries. Alternatively, \textit{de novo} design approaches \cite{bohacek_growmol_nodate,spiegel_autogrow4_2020} have re-emerged with the advent of deep learning-based generative methods adapted to build molecules with enhanced affinity for a target structure of interest. A variety of generative approaches have been explored, including Conditional Variational Autoencoder (cVAE) \cite{ragoza_generating_nodate}, flow-based models \cite{shi_graphaf_2020, jiang_pocketflow_2024, 10.5555/3692070.3693767, cremer_flowr_2025}, diffusion models \cite{guan2023d, schneuing_structure-based_2024, guan_decompdiff_2024}, and Generative Pretrained Transformers (GPTs) \cite{wu_tamgen_2024}.

\begin{figure*}[h]
    \centering
    \includegraphics[width=0.9\linewidth]{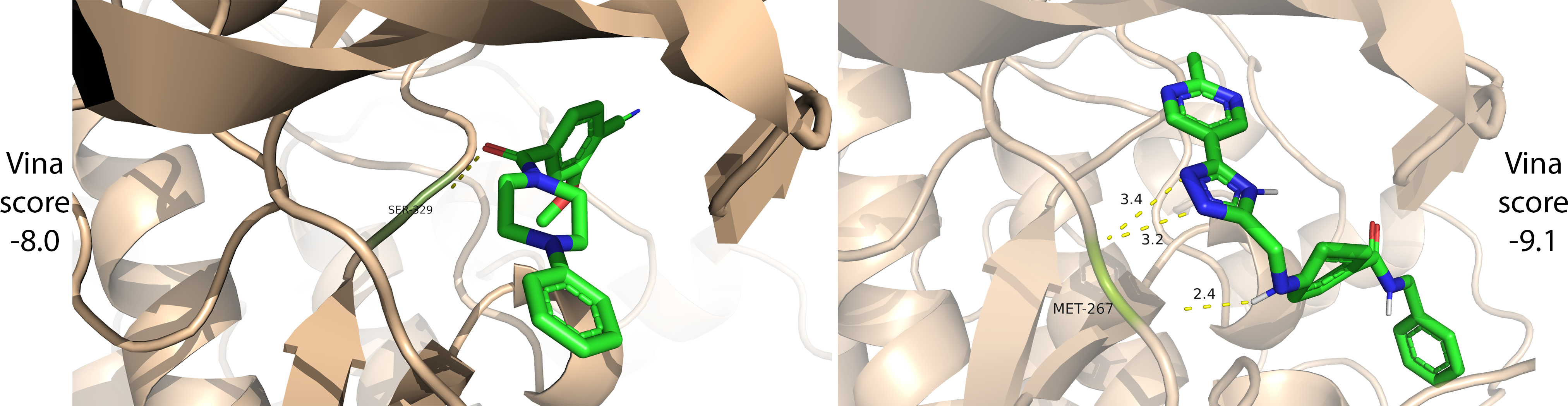}
    \caption{Docked binding poses of generated molecules without (left) and with (right) BioAssay context. With BioAssay context, ChatGPT 4o generates a molecule with three hydrogen bonds to the GRK4 pocket residue MET-267, improving the docking score.}
    \label{fig:proof_of_concept}
\end{figure*}

\section{Assay2Mol framework}
\subsection{Motivating example}
Before designing a general algorithm, we begin with a proof of concept to assess whether an LLM generates relevant molecules and how BioAssay context affects generation. Using the UniProt \citep{uniprot_2023} protein description of GRK4 (UniProt P32298; PDB 4YHJ) as input, we prompt ChatGPT 4o to generate five molecules. Next, we use the same protein description to search for related BioAssays (see Section 3.3) and retrieve four BioAssays related to GRK4: 775998\footnote{\url{https://pubchem.ncbi.nlm.nih.gov/bioassay/775998}} (human GRK2), 1315729 (human GRK2), 775996 (human GRK5), 1315749 (bovine GRK2).
We generate another five molecules, this time providing the retrieved BioAssay data, including the experimental tables, as additional context to ChatGPT 4o. 
 The average AutoDock Vina \cite{doi:10.1021/acs.jcim.1c00203} score for the five molecules without BioAssay context is -7.44 (expressed in units of kcal/mol). The average Vina Dock score for the four molecules with BioAssay context is -8.48 (one was invalid). Lower scores reflect better structural complementarity between the generated molecule and the protein target. Docked poses for the top scoring molecule from each group are shown in Figure \ref{fig:proof_of_concept}.
This small pilot study supports the proposition that the BioAssay context can improve molecule generation and motivates a full exploration of that problem setting.

\subsection{Problem definition}
Given a text description $p$ for a target protein or phenotype, we want to retrieve the most relevant BioAssays $b$. Then, based on the experimental results associated with the BioAssays $b$, we want to generate molecules that produce the desired target response or activity.
Below we use "query protein", though it could be a target protein or phenotype.

\subsection{BioAssay retrieval}

The BioAssay retrieval stage is similar to Retrieval Augmented Generation (RAG) \citep{lewis_retrieval-augmented_2021}. For a query description, we extract keywords with an LLM and obtain a protein description embedding $\mathbf{p} \in \mathbb{R}^d$. We use the OpenAI text embedding tool \citep{neelakantan_text_2022} and obtain an embedding for BioAssay record $i$ in json format, recorded as $\mathbf{b}_i \in \mathbb{R}^d$ \citep{douze_faiss_2025}. We use cosine similarity to calculate the similarity between the query protein description and the set of available PubChem BioAssays and then select the top-$k$ related BioAssays:

\begin{equation}
\begin{aligned}
  &\mathcal{I}_k = \text{arg top-}k \left\{ \frac{\mathbf{p} \cdot \mathbf{b}_i}{\|\mathbf{p}\| \|\mathbf{b}_i\|} : i = 1, 2, \dots, N \right\} \\
&\{\mathbf{b}_i : i \in \mathcal{I}_k\}.  
\end{aligned}
\end{equation}
In contrast to RAG, we do not use the retrieved BioAssays as context directly. Instead, we download data tables of these BioAssays based on their Assay ID (AID) and perform further filtering:
\begin{itemize}
    \item To ensure fair comparisons, BioAssays directly involving the query protein, identified by matching UniProt IDs, are excluded.
    \item Many BioAssays are derived from literature and involve only a single or few molecules. We prioritize BioAssays with larger data tables using a filter threshold $min\_mol\_num$, which removes BioAssays with fewer molecules tested.
    \item More shots (examples) will usually lead to better performance with in-context learning. However, the context length of an LLM is limited. Thus, we set $max\_assay\_num$ to limit the number of BioAssays retrieved.
    \item Sometimes the query protein has no relevant BioAssays. In such cases, the retrieved BioAssays with the highest cosine similarity would be uninformative. To further assess relevance of the retrieved BioAssays, we also use an LLM to determine whether the retrieved BioAssays are relevant to the query protein.
\end{itemize}

\subsection{Counterscreen BioAssay}
BioAssay records are sometimes grouped by project. Projects contain independent records for primary screens (often high-throughput screens) and confirmatory screens (secondary re-testing of hits from the primary screen). Of particular interest are counterscreens, which are designed to detect false positives such as pan-assay interference compounds (PAINS) \citep{baell_seven_2018} or assess hit specificity by testing for activity on an off-target or "anti-target," an undesirable, sometimes related target. Active molecules in counterscreens are  undesirable and should be avoided or used as negative training instances.
Therefore, we use the LLM to summarize the retrieved BioAssays and identify whether the BioAssay record represents a counterscreen assay (Appendix \ref{BioAssay summarization Prompt}). 

\subsection{Layered contextual analysis}
The LLM sequentially processes the set of relevant BioAssay records to build an input prompt for molecule generation. The workflow processes each BioAssay in three steps:
\begin{enumerate}
    \item \textbf{Summarization of BioAssay findings:}
The LLM generates a concise summary that captures the purpose, methodology, and key results of each BioAssay. Additionally, the LLM states the apparent relationship between the BioAssay and the query protein, considering how the BioAssay's findings may inform the design of new molecules. If the BioAssay is a counterscreen, its active molecules should be avoided.
\item \textbf{Presentation of tabular experimental data:} For each BioAssay, the experimental data are presented in a table describing the SMILES notation of each molecule and its activity result (Active, Unspecified, or Inactive). Most experimental data tables also include measured pharmacodynamic parameters expressed using standard types (e.g., $IC_{50}$, $K_i$, $K_d$, percent inhibition), relations (e.g., $<, =, >$), values, and units (e.g., $\mu M$, \%).
\item \textbf{Molecule selection:}
If active molecules are identified in the data table, we next check the number of actives.
If the number of actives exceeds $N_{mol}$, we randomly sample $N_{mol}$. Otherwise, we list all active molecules.
To maintain class balance, we randomly sample $N_{mol}$ molecules from the combined unspecified and inactive categories.

If there are no active molecules, we include all molecules unless they exceed $2 \cdot N_\text{mol}$, in which case we instead randomly sample $2 \cdot N_\text{mol}$. Accordingly, we increase $min\_mol\_num$ to $2 \cdot N_\text{mol}$ to account for the lack of actives.
\end{enumerate}
An example of the BioAssays summarization is in Appendix \ref{summarization example}.

\subsection{Molecule generation}
Given the context of the query protein description and BioAssay summaries and data tables, Assay2Mol uses the LLM to generate molecules in batches of 10. Details are in the prompt in Appendix \ref{Generation Prompt}. The full Assay2Mol workflow is shown in Figure \ref{fig:workflow}.

\section{Experiments}

We evaluate Assay2Mol in two settings. First, we compare Assay2Mol with SBDD methods for generating candidate protein-binding molecules.
Second, we examine Assay2Mol's ability to manage multiple objectives by generating molecules that bind a query protein and avoid cardiotoxicity.

\begin{table*}[!ht]
\centering
\caption{Experimental results on the CrossDocked dataset. The best results are shown in \textbf{bold}, and the second-best results are \underline{underlined}.}
\resizebox{\textwidth}{!}{
\begin{tabular}{lcccccccccccc} 
\toprule
\multirow{2}{*}{Model} & \multicolumn{2}{c}{Vina Dock ($\downarrow$)} & \multicolumn{2}{c}{High Affinity ($\uparrow$)} & \multicolumn{2}{c}{QED ($\uparrow$)} & \multicolumn{2}{c}{SA ($\uparrow$)} & \multicolumn{2}{c}{Diversity ($\uparrow$)} &
\multicolumn{2}{c}{Size} \\ 
\cmidrule(lr){2-3} \cmidrule(lr){4-5} \cmidrule(lr){6-7} \cmidrule(lr){8-9} \cmidrule(lr){10-11} \cmidrule(lr){12-13} 
& Avg. & Med. & Avg. & Med. & Avg. & Med. & Avg. & Med. & Avg. & Med. & Avg. & Med.  \\ 
\midrule
Reference   &-7.117  & -6.905 & - & -& 0.476 & 0.468 & 0.728 & 0.740 & - & - &22.75& 21.50   \\ 
FDA drugs & -7.027 & -7.169 & 0.440 & 0.380 & 0.567 & 0.564 & 0.760 & 0.758 & \multicolumn{2}{c}{0.792} &  \multicolumn{2}{c}{24.47} \\
ZINC 30 atoms & -7.855 & -8.088 & 0.607 & 0.735 & 0.576 & 0.577 & 0.737 & 0.736 & \multicolumn{2}{c}{0.655} & \multicolumn{2}{c}{30.0}\\
\hline
CVAE & -6.114 & -6.118 & 0.103& 0.026& 0.390 &0.419 & 0.591 & 0.580 & 0.655&0.666 & 19.97&20.19  \\ 
AR         & -6.751 & -6.707 & 0.459 & 0.340& 0.505 & 0.499 & 0.635 & 0.634 & 0.698 & 0.703 & 17.78 & 17.54     \\ 
Pocket2Mol   & -7.200 &  -6.815 & 0.601& 0.593 & 0.574 & 0.579 & 0.754& 0.760 & 0.741& 0.781 & 17.84& 16.53   \\  
TamGen & -7.475 & -7.775 & 0.526& \underline{0.645}& 0.559 & 0.559 & 0.771 & 0.759 & \underline{0.747} & \underline{0.745} & 23.13 & 23.29   \\ 
TargetDiff   & -7.788 & \underline{-7.964}&  \textbf{0.683}& 0.634 & 0.474 & 0.485 & 0.584& 0.571 & 0.717 & 0.714 & 24.44 & 24.64    \\ 
\hline
Gemma-3-27B & -7.050 & -7.024 & 0.416 & 0.281 & 0.700 & 0.711 & \underline{0.860} & \underline{0.868} & 0.757 & 0.765 & 19.34 & 18.94  \\
GPT 4o & -7.198 & -7.257 & 0.432 & 0.294& \textbf{0.789} & \textbf{0.803} & \textbf{0.870} & \textbf{0.878} & \underline{0.767} & \underline{0.767} & 19.70 & 19.64 \\
DeepSeekV3 & -7.230 & -7.170 & 0.443 & 0.241 & \underline{0.743} & \underline{0.756} & 0.855 & 0.867 & \textbf{0.771} & \textbf{0.772} & 18.96 & 19.00 \\
\hline
Assay2Mol (Gemma-3-27B) & \textbf{-8.064} & \textbf{-8.280} & \underline{0.610} & \textbf{0.732} & 0.585 & 0.606 & 0.821 & 0.834 & 0.742 & 0.611 & 26.59 & 26.65 \\
Assay2Mol (GPT 4o) & -7.796 & -7.881 & 0.548 & 0.576 & 0.600 & 0.630 & 0.790 & 0.801  & 0.542 & 0.547 & 25.90 & 25.59  \\ 
Assay2Mol (DeepSeekV3) & \underline{-7.861} & -7.936 & 0.557 & 0.634 &  0.616 & 0.647 & 0.813 & 0.820 & 0.593 & 0.608 & 24.46 & 24.26  \\
Assay2Mol (DeepSeekV3 <30\%) & -7.826 & -7.925 & 0.562 & 0.673 & 0.594 & 0.619 & 0.814 & 0.820 & 0.609 & 0.628 & 24.49 & 24.73 \\

\bottomrule
\end{tabular}
}
\label{tab:crossdock_result}
\end{table*}

\begin{table*}[htb]
    \centering
\caption{Average improvement over randomly sampled FDA drugs grouped by LLM-estimated relevance of the retrieved BioAssays. The value represents the increase in the docking score, measured in kcal/mol.}
\resizebox{\textwidth}{!}{
\begin{tabular}{lccccccccccc}
\toprule
\multirow{2}{*}{Model} & \multicolumn{2}{c}{High (39\%)} & \multicolumn{2}{c}{Medium (42\%)} & \multicolumn{2}{c}{Low (7\%)} & \multicolumn{2}{c}{No (12\%)} & \multicolumn{2}{c}{Overall}  \\
\cmidrule(lr){2-3} \cmidrule(lr){4-5} \cmidrule(lr){6-7} \cmidrule(lr){8-9} \cmidrule(lr){10-11} 
& Avg. & Med. & Avg. & Med. & Avg. & Med. & Avg. & Med. & Avg. & Med.\\
\hline
TargetDiff & 0.838 & 0.802 & 0.701 & 0.777 & 0.669 & 0.696 & 0.771 & 1.052 & 0.761 & 0.796 \\
\hline
Gemma-3-27B & 0.196 & 0.170 & -0.050 & -0.145 & 0.197 & 0.293 & -0.390 & -0.535 & 0.023 & 0.034 \\
GPT 4o & 0.331 & 0.228 & 0.116  & 0.118 & 0.396 & 0.302 & -0.289 & -0.122 & 0.171 & 0.130 \\
DeepSeekV3 & 0.379 & 0.311 & 0.079 & 0.070 & 0.429 & 0.300 & -0.067 & -0.098 & 0.203 & 0.159 \\
\hline
Assay2Mol (Gemma-3-27B) & 1.277 & 1.124  & 1.037& 1.121& 0.535& 0.770&  0.554& 0.606 & 1.037 & 1.069 \\
Assay2Mol (GPT 4o) & 1.061 & 1.046 & 0.732 & 0.741 & 0.223 & 0.517 & 0.269& 0.151 & 0.769 & 0.777 \\
Assay2Mol (DeepSeekV3) & 1.042 & 0.634 & 0.842 & 0.921 & 0.599& 0.579 & 0.267 & 0.273 & 0.834 & 0.849 \\

\midrule
\end{tabular}
}
    \label{tab:normalized_table}
\end{table*}

\subsection{Generating binders for target proteins}
\textbf{Dataset.} CrossDocked2020 (CrossDocked for short) is a common dataset for SBDD \citep{francoeur_CrossDocked2020_2020} that allows us to assess how BioAssay context compares to protein structure context for generating candidate protein binders. Previous methods refined the original 22.5 million docked protein binding complexes by isolating those with poses $<$ 1 \r{A} RMSD from native (crystallographic poses) and sequence identities $<$ 30\% from the original dataset. They used 100,000 complexes for training and 100 novel complexes as references for testing. Assay2Mol does not require additional training. We select 100 complexes from the training set to develop our pipeline and then evaluate on the test set. As input prompts for protein targets, we use the descriptions returned from PubChem from queries of the PDB ID of each protein. For proteins whose PDB ID cannot be found in PubChem, we use the UniProt mapping tool to convert the PDB ID into the UniProt ID, which is then used to query the PubChem protein webpage. When this approach fails, we manually collect information about the protein from the literature listed on its PDB homepage \citep{burley_rcsb_2023}.

\textbf{Methods.} Before evaluating Assay2Mol on CrossDocked, we selected its hyperparameters using nine proteins from the CrossDocked training set. We set $max\_assay\_num$ to 10,  $N_{mol}$ to 8, and $max\_mol\_size$ to 45. We filter out molecules greater than $max\_mol\_size$ from the input context in the BioAssay data table, which helps control the size of the generated molecules. 
We test Assay2Mol with three LLMs: Gemma-3-27B \citep{team_gemma_2025}, DeepSeekV3 \citep{deepseek-ai_deepseek-v3_2024}, and GPT 4o \citep{openai_gpt-4o_2024}. Details are in Appendix \ref{sec:hyperparameter}. To assess whether data leakage affects the results, we run Assay2Mol with an additional filter with DeepSeekV3. After retrieving each BioAssay, we compute the sequence identity between its associated protein and the query protein with MMseqs2 \citep{steinegger_mmseqs2_2017}. If the sequence identity exceeds 30\%, the BioAssay will be discarded and replaced by the next candidate until $max\_assay\_num$ BioAssays are collected. This result is denoted as Assay2Mol (DeepSeekV3 <30\%) in Table \ref{tab:crossdock_result}.

We compare Assay2Mol against the following SBDD methods: CVAE \citep{ragoza_generating_nodate}, AR \citep{luo_3d_2022}, Pocket2Mol \citep{peng_pocket2mol_2022}, GraphBP \citep{liu_generating_2022}, TamGen \citep{wu_tamgen_2024}, and TargetDiff \citep{guan2023d}. Among these, TamGen generates SMILES, whereas the other methods generate 3D molecule conformations.
We obtain the previously generated TamGen results from its repository. Docking scripts and other methods' results come from the TargetDiff repository. We use the existing generated molecules from files linked in these repositories directly for evaluation. When rerunning the docking and evaluation scripts, some results may differ from those reported in previous papers due to differences in computing environments. We prioritized comparing to these SBDD methods that had readily available generated molecules or scripts.

The Gemma-3-27B, GPT 4o, and DeepSeekV3 methods are an Assay2Mol ablation that assesses how much of the molecule generation capability comes from the BioAssay context.
These methods generate molecules with an LLM using the protein description alone, as shown in Appendix \ref{ablation_prompt}.

\textbf{Metric.} We use \textbf{Vina Dock} \citep{doi:10.1021/acs.jcim.1c00203} to score the binding complementarity of a molecule, or the strength of interaction, with a protein target. \textbf{High Affinity} indicates the percentage of generated molecules that outperform the reference molecules in \textbf{Vina Dock}. Quantitative Estimate of Drug-likeness (\textbf{QED}) is a metric that combines multiple molecular properties (e.g., molecular weight, logP, hydrogen bond donors) into a single value between 0 and 1, with higher values indicating more drug-like compounds \citep{bickerton_quantifying_2012}. Synthetic Accessibility (\textbf{SA}) score estimates synthetic feasibility based on fragment contributions observed in known molecules and structural complexity penalties \citep{ertl_estimation_2009}. We use the normalized SA score between 0 and 1, where 0 is most difficult to synthesize.
QED and SA are computed with RDKit \citep{Landrum2016RDKit2016_09_4}. \textbf{Diversity} is quantified as the average pairwise Tanimoto distance between Morgan fingerprint of the generated molecules.
The Vina score is correlated with the number of atoms in the molecule \citep{weller_structure-based_2024}, so we also track the \textbf{Molecule Size} as the number of heavy atoms. In order to demonstrate the influence of molecule size on docking score, we randomly sample 100 molecules with 30 heavy atoms from ZINC20 \citep{doi:10.1021/acs.jcim.0c00675} and add this baseline to Table \ref{tab:crossdock_result}. We also discuss the molecule \textbf{validity} rate and \textbf{price} for different LLMs in Appendix \ref{sec:llm_validity}.

Metrics are calculated for each protein target. First, we compute the average metrics of the generated molecules for each corresponding protein. Then, we calculate the mean and median scores across all 100 proteins. As an additional baseline that can highlight protein-specific biases in docking scores, we randomly sample 100 FDA-approved drugs.

\textbf{Results.}
Most of the existing SBDD methods such as CVAE, AR, and Pocket2Mol do not perform well on average.
This may be partially due to the size of the molecules they generate.
The random sample of large, irrelevant ZINC molecules with 30 heavy atoms shows that they produce better docking scores than many of the generative models.

All versions of Assay2Mol consistently outperform the best SBDD method, TargetDiff, in average docking scores (Table \ref{tab:crossdock_result}). Assay2Mol (Gemma-3-27B) produces better docking scores than the other Assay2Mol variants, though it generates larger molecules, making it appealing as a locally-run open weights model that can process BioAssay context. The performance of Assay2Mol (DeepSeekV3) remains stable after removing proteins with more than 30\% sequence identity. This indicates that the observed performance is not attributable to potential data leakage from the most closely related proteins, but rather reflects the model's ability to generalize. 

Beyond improved docking scores, Assay2Mol generates molecules with relatively high synthetic accessibility and QED scores, benefiting from LLMs' molecule generation capability. However, the GPT 4o and DeepSeekV3 versions perform poorly in terms of molecular diversity. We find that LLMs tend to generate similar molecules within a group when context molecules are provided. In extreme cases in our preliminary testing that generated 100 molecules per batch, the model incrementally added a single carbon atom to the molecular backbone each time. The current setting of 10 molecules per batch alleviates this issue and helps balance computational costs and molecule diversity.
It is surprising that the three LLMs in the Assay2Mol ablation can generate high-quality molecules in a zero-shot setting, outperforming some SBDD models. These molecules also exhibit desirable drug-likeness and synthetic accessibility characteristics (QED, SA). However, without guidance from the BioAssay context, the generated molecules tend to be smaller in size and exhibit less favorable docking scores than Assay2Mol.
We also compute the similarity between generated molecules and context molecules to demonstrate that the LLMs learn from context rather than merely making minor modifications to existing molecules. More details are shown in Appendix \ref{similarity_analysis}.

A more detailed examination of the mouse protein TFPI provides context to Assay2Mol's performance on the CrossDocked dataset (Appendix \ref{TFPI discussion}). It reveals that not all BioAssays with similar embeddings to the target protein query are biologically relevant. This is why we use an LLM to estimate the relevance of the retrieved BioAssays to the query protein. For each query protein, we analyze the top 10 BioAssays after filtering. We run GPT 4o and DeepSeek-V3 and then aggregate their results. We define $x$ as (relevant BioAssays)/(total BioAssays). This categorizes the proteins into four groups: high relevance ($x \ge 0.7$), medium relevance ($0.4 < x < 0.7$), low relevance ($0.1 < x \leq 0.4$), and no relevance ($x \leq 0.1$). The results of different groups are shown in Figure \ref{fig:group}. As expected, Assay2Mol performs worst on the no relevance group compared with TargetDiff, and it is consistent with our small-scale evaluation of low-similarity BioAssays (Figure \ref{fig:dissimilar_plot}).

To check the accuracy of the LLMs' relevance evaluation, we manually evaluate the BioAssays retrieved for 25 targets and find that GPT 4o is fairly accurate in its relevance assessment but DeepSeek-V3 struggles (Table~\ref{tab:relevance-review}). DeepSeek-V3 incorrectly assesses the relevance for all ten BioAssays for six different targets. 
We perform the manual assessment by reading through the assay description and abstract for each of the 10 retrieved AIDs for each target protein (based on PDB code) and confirming whether the assay readout would correspond with a biochemical interaction of the tested compounds with a protein of the same family as the target protein. Also, where applicable we also confirmed the desired functional activity, inhibition or activation. In cases where it was not clear, the AID was labeled as not relevant.

Because docking score distributions differ across proteins, directly averaging scores as in Table \ref{tab:crossdock_result} may fail to capture meaningful improvements over baseline methods. To account for protein-specific effects, we dock 100 FDA-approved drugs to the 100 CrossDocked test proteins to establish a baseline score distribution. We then compute the improvement in docking score for each generated molecule with respect to these protein-specific background distributions. The improvement of Assay2Mol decreases as BioAssay relevance decreases, while TargetDiff demonstrates relatively uniform performance across all groups, consistent with its reliance solely on protein structure (Table \ref{tab:normalized_table}). GPT 4o outperforms Assay2Mol (GPT 4o) in the low relevance group, suggesting that the inclusion of irrelevant BioAssay context may mislead the LLM for molecule generation. LLMs without context perform poorly in the no relevance group, indicating that these proteins are possibly understudied and underrepresented in existing databases and their pretraining data. SBDD is a more suitable strategy in such cases. These observations reinforce the validity of the Assay2Mol concept, suggesting that the LLM benefits from assay context to guide molecule generation. Also, most of the proteins in the CrossDocked test set fall into the "High" and "Medium" groups (81\%), indicating Assay2Mol's practical utility for potential targets of interest.

\begin{figure*}[t]
    \centering
    \includegraphics[width=\linewidth]{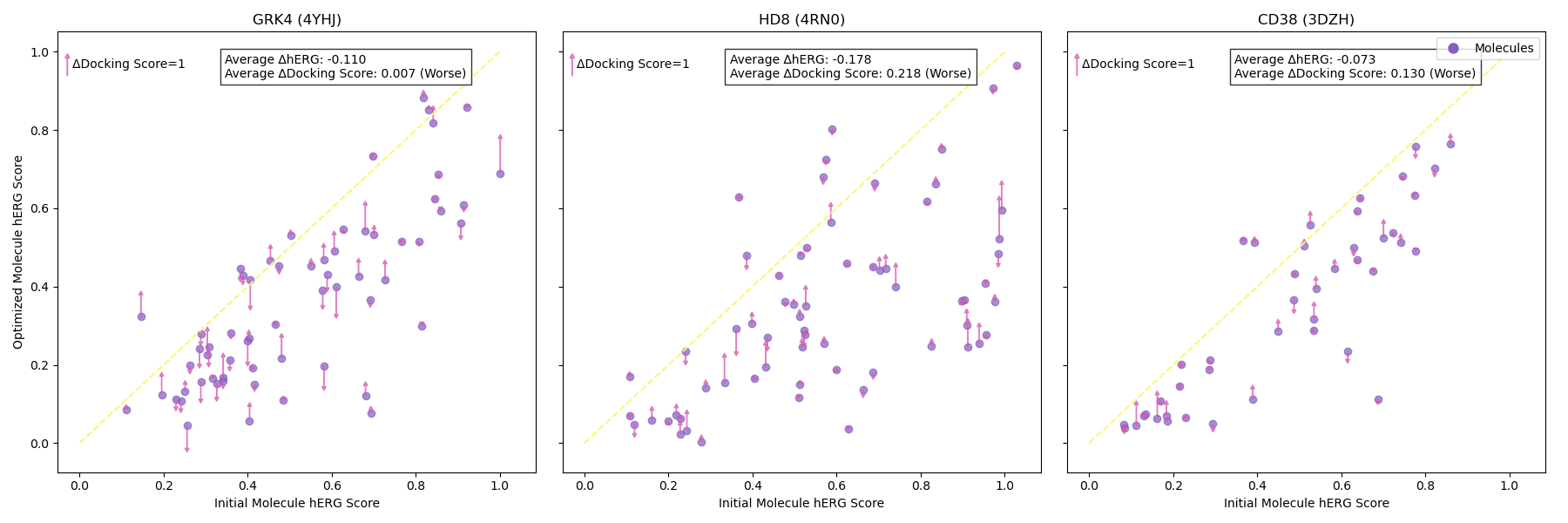}
    \caption{Change in predicted hERG score and docking score between initial and optimized molecules for three proteins. The up arrow indicates the docking score decreases and the down arrow indicates it increases. The length of the arrow (top-left) serves as a scale bar, representing an increase of 1 score unit (kcal/mol) from Vina Dock.}
    \label{fig:hERG}
\end{figure*}

\subsection{Specificity and counterscreen with hERG}
KCNH2, also known as hERG, is a voltage-gated potassium ion channel that plays a crucial role in cardiac repolarization.
Blocking hERG channels with drugs can lead to prolonged QT intervals, potentially causing severe cardiac arrhythmias or sudden death. It has been one of the most frequent adverse side effects leading to drug failure \citep{sanguinetti2006herg}. As a result, evaluating a molecule’s interaction with hERG is a critical step in drug development to ensure safety. In the PubChem database, there are many BioAssays using hERG as a counterscreen \citep{GARRIDO2020112290}.
We ask:
\begin{enumerate}[nosep]
    \item Can Assay2Mol accurately interpret the counterscreen context to reduce the generated molecule's affinity for hERG?
    \item Can the generated molecules retain high affinity to the original target protein?
\end{enumerate}
We selected proteins GRK4 (PDB: 4YHJ), HD8 (PDB: 4RN0), and CD38 (PDB: 3DZH) from the CrossDocked test set. These proteins serve as potential targets for cancer treatments or antibiotics. Since the corresponding molecules may potentially be developed as drugs taken by humans, it is crucial to evaluate their potential hERG-related interactions to assess safety. After we generate molecules for these proteins using the same methods as in the CrossDocked evaluation, we use the description of hERG\footnote{\url{https://pubchem.ncbi.nlm.nih.gov/gene/3757}
} to search for related BioAssays, in the same manner used to formulate the context described in Section 3. After we prepare the whole context, we append generated molecules from the previously selected proteins and ask the LLM to optimize these molecules to reduce binding affinity toward hERG. The prompt is shown in Appendix \ref{hERG prompt}. We use ADMETlab 3.0 \citep{10.1093/nar/gkae236} to predict the hERG score. To avoid circular reasoning, we verify that no molecules in the hERG generation context were present in the ADMETlab 3.0 hERG training set.

We examine the shift in Vina docking scores and predicted hERG scores for the original generated molecules versus those optimized to minimize hERG interaction (Figure \ref{fig:hERG}) The average hERG score of molecules for 4YHJ decreases from 0.503 before optimization to 0.393 after optimization. For comparison, the average hERG score of 2,965 FDA-approved drugs is 0.284. This reduction in hERG score indicates lower predicted cardiotoxicity, making the generated molecules more suitable for further study, while docking scores remain largely unaffected, demonstrating that in-context learning with a counterscreen enhances specificity without compromising affinity.

\section{Discussion and conclusions}
Our initial version of Assay2Mol can successfully query PubChem with text descriptions, retrieve relevant BioAssays, and generate molecules based on the text and screening data from those relevant BioAssays.
Assay2Mol shows how to use LLMs to make better use of decades' worth of valuable, unstructured chemical screening data in PubChem.
Our approach generates molecules with docking scores comparable to SBDD methods and advances controllable natural language-driven molecule design.
There are many opportunities to expand on the core Assay2Mol framework, making it more robust and building new capabilities to make it more relevant for the difficult multi-property optimization required for actual drug discovery campaigns \citep{van_den_broek_search_2025}.
For example, we observed that the initial embedding-based query and BioAssay similarity calculations are not sophisticated enough to capture the complexity of biological regulation in pathways (Appendix \ref{TFPI discussion}) along with other limitations in Section \ref{limitations}.

\section{Limitations}
\label{limitations}
There are several opportunities to improve how LLMs are used within Assay2Mol.
Having LLMs directly assess the relevance of the retrieved BioAssay text guards against many irrelevant matches but is imperfect (Table~\ref{tab:relevance-review}).
We continue to explore methods to enhance LLM interpretation of the desired activity of generated molecules with respect to target function (i.e., activation, inhibition, allosteric regulation).
A related limitation is that the current version of Assay2Mol cannot properly process conditional text queries, such as molecules that inhibit proteins A, B, and C but not D and E.
This limitation is again related to the initial embedding-based similarity calculations, which could possibly be addressed with enhancement to the Assay2Mol relevance processing step.
Our hERG example shows that Assay2Mol can be used for conditional molecule generation, for example inhibiting GRK4 but not hERG, if these steps are run sequentially.
Furthermore, there is potential for improvement in both construction of text prompts and the LLMs used within the Assay2Mol framework.
Our BioAssay relevance assessment evaluation showed that GPT 4o matched manual assessments much more closely than DeepSeek-V3 (Table~\ref{tab:relevance-review}), so the choice of LLM can impact the overall results.
Finally, we have not yet evaluated the sensitivity of Assay2Mol to different LLM prompting strategies or optimized the prompts.

LLMs are pre-trained on a large-scale text corpus, which includes a substantial number of molecules. As a result, when generating molecules, LLMs may not always create new molecules. Instead, they tend to "retrieve" molecules or recombine patterns from molecules they have trained on. Repurposing existing molecules for new targets or modifying existing molecules means that they tend to exist in certain databases, be available for purchase, or be more feasible to synthesize chemically. However, for researchers whose primary goal is to push beyond known chemical space, this characteristic might be regarded as a limitation rather than an advantage.

Most of the LLM-based steps in Assay2Mol are implemented using both closed and open weights LLMs with the goal of having a fully open weight version of Assay2Mol.
However, currently the BioAssay embeddings are generated solely using the OpenAI text embedding API.
An alternative implementation and evaluation with open weights embedding models remains for future work.
Even the open weights LLMs Assay2Mol uses are not fully open source and do not have their training data available.
This lack of training data and what biochemistry data are included makes it challenging to fully interpret the Assay2Mol-generated molecules and their limitations.
If it is generating novel molecules as opposed to retrieving molecules from the LLM training set, general caveats about generative molecular design apply to those outputs \citep{walters_generative_2024}.

The evaluations of the Assay2Mol generated molecules rely entirely on other computational assessments of molecule quality. These are not a substitute for actual wet lab assays. Vina Dock energies are not true binding affinities, and its scoring function has known biases and limitations \citep{xu_systematic_2022}. For instance, we found that randomly sampled ZINC molecules with 30 heavy atoms produced docking scores comparable to Assay2Mol (Table \ref{tab:crossdock_result}), illustrating the known relationship between docking score and molecule size \citep{weller_structure-based_2024}.
 In addition, the hERG scores are computed with an existing regressor that has reasonably good but imperfect performance \citep{10.1093/nar/gkae236}.

\section*{Software availability}
The code is available at \url{https://github.com/gitter-lab/Assay2Mol} under the MIT License and archived at \url{https://doi.org/10.5281/zenodo.15871304}. The datasets are available at \url{https://doi.org/10.5281/zenodo.15867214} under the CC BY 4.0 license.

\section*{Acknowledgments}
This research was supported by National Institutes of Health award R01GM135631.

\bibliography{custom}

\clearpage

\appendix

\section{Appendix}
\label{sec:appendix}

\begin{figure*}
    \centering
    \includegraphics[width=1.0\linewidth]{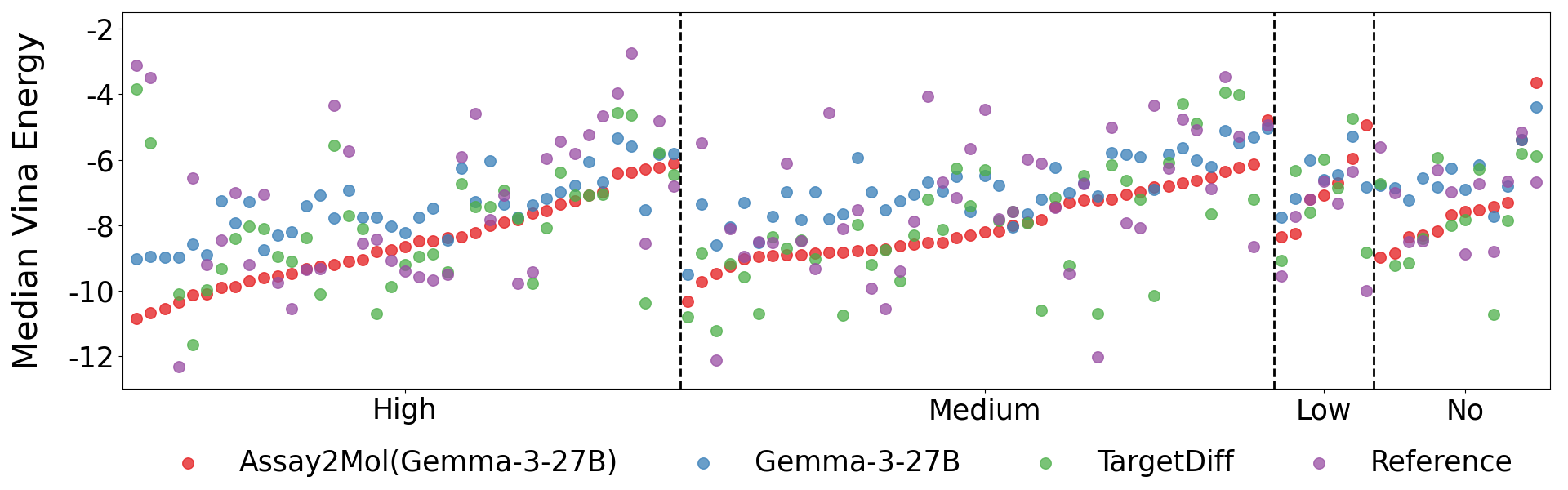}
    \centering
    \caption{Distribution of docking scores of different relevance level groups. }
    \label{fig:group}

\end{figure*}

\subsection{Potential risks}
Like many other molecule generation tools, Assay2Mol could be considered a dual-use technology.
Its primary intended application is in early stage drug discovery as a strategy to sift through unstructured experimental BioAssay descriptions in PubChem.
However, BioAssays in PubChem used for counterscreening molecules measure toxicity, so it could be possible to use Assay2Mol to generate toxic molecules.
Existing generative models can also be trained or guided with this public toxicity data, so this is a risk of a broad class of models not Assay2Mol specifically.

Assay2Mol generates molecules tailored to activity against the target protein or phenotype.
Because the prompts are constructed in a structured manner using BioAssay context, it is less likely to be able to generate some classes of harmful molecules like explosives compared to generative models that take free text descriptions of molecule properties as input.
LLMs perform the actual molecule generation, and we have not formally studied how the BioAssay context expands or reduces risk relative to the baseline LLMs.
We anticipate that Assay2Mol's structured prompts restrict the types of molecules the LLMs can generate, and some of the generated molecules are actually retrieved existing molecules.

\subsection{Hyperparameter and model selection}
\label{sec:hyperparameter}
For the Assay2Mol hyperparameter selection, we choose nine proteins from the randomly sampled 100 proteins from the training set: 1RQP, 3ANT, 4A9S, 4I29, 4XE6, 4Y2R, 5ACC, 5AI4, 5PNX. We manually inspect their retrieved BioAssays and make sure they are relevant. We also include the reference molecule and label it as "CrossDocked". Assay2Mol invloves three hyperparameters: $max\_assay\_num$, $N_{mol}$, and $max\_mol\_size$, making it time-consuming to perform grid-search. To simplify the process, we fix two hyperparameters when optimizing the remaining one.

\begin{figure}[htbp]
    \centering
    \includegraphics[width=1.0\linewidth]{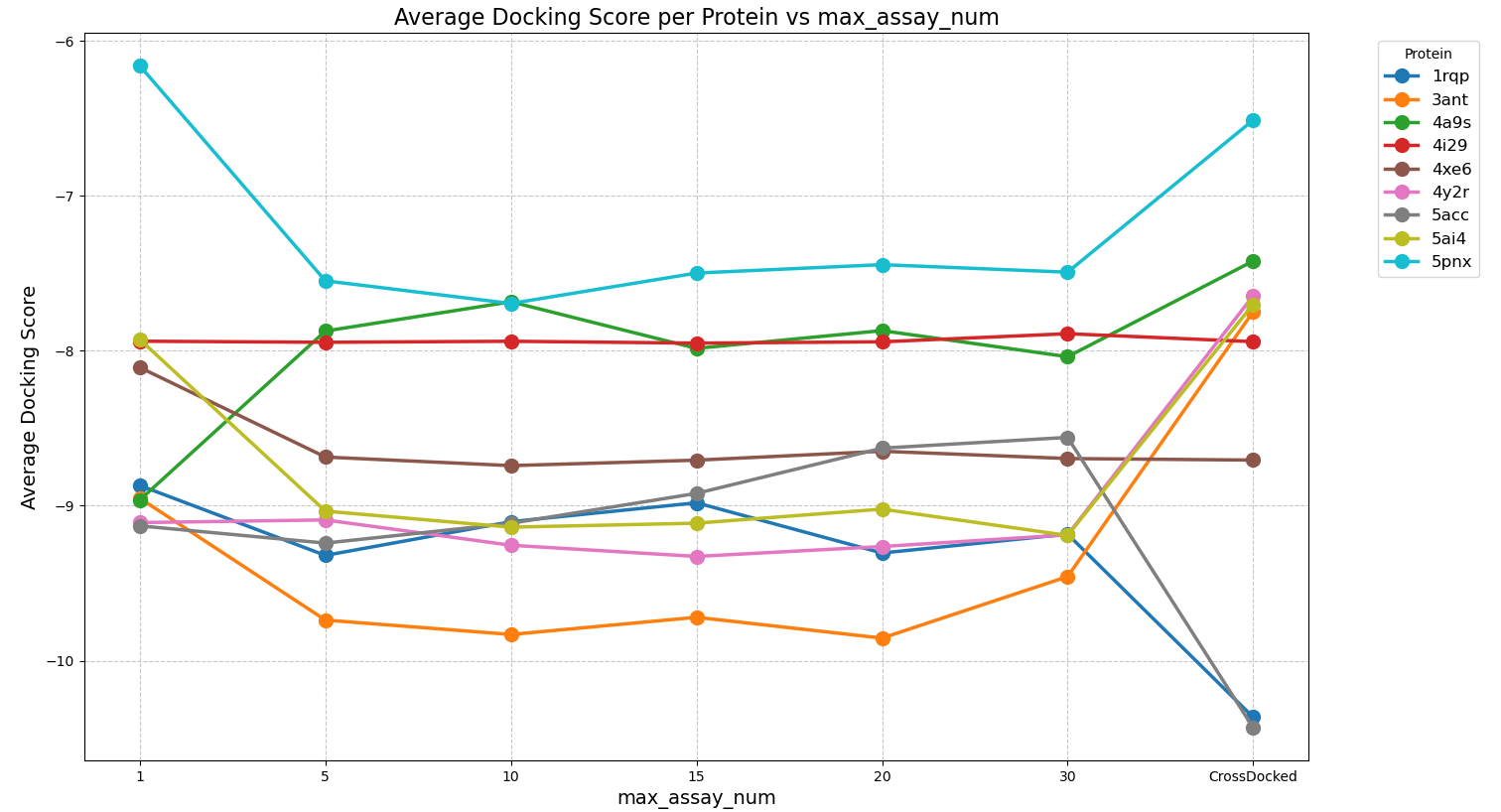}
    \caption{Average docking score under different $max\_assay\_num$}
    \label{fig:max_assay_num}
\end{figure}

\begin{figure}[htbp]
    \centering
    \includegraphics[width=1.0\linewidth]{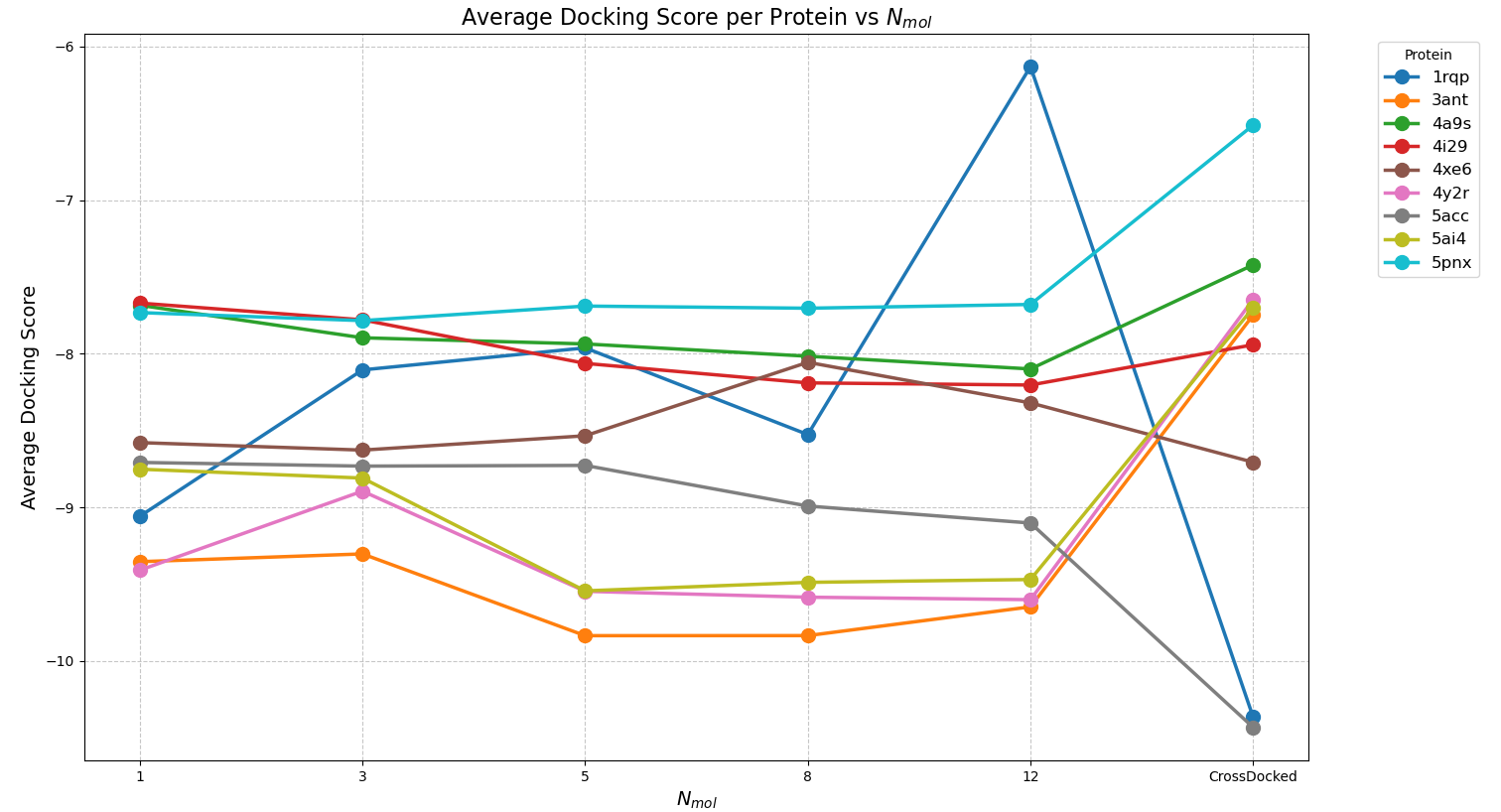}
    \caption{Average docking score under different $N_{mol}$}
    \label{fig:N_mol}
\end{figure}

First, we set $N_{mol}$ to 5, impose no limit on $max\_mol\_size$, and vary $max\_assay\_num$ across [1, 5, 10, 15, 20, 30]. According to Figure \ref{fig:max_assay_num}, we observe that $max\_assay\_num=10$ yields strong performance. More than 10 BioAssays leads to longer context and increases computational cost without significant gains.

\begin{figure}[htbp]
    \centering
    \includegraphics[width=1.0\linewidth]{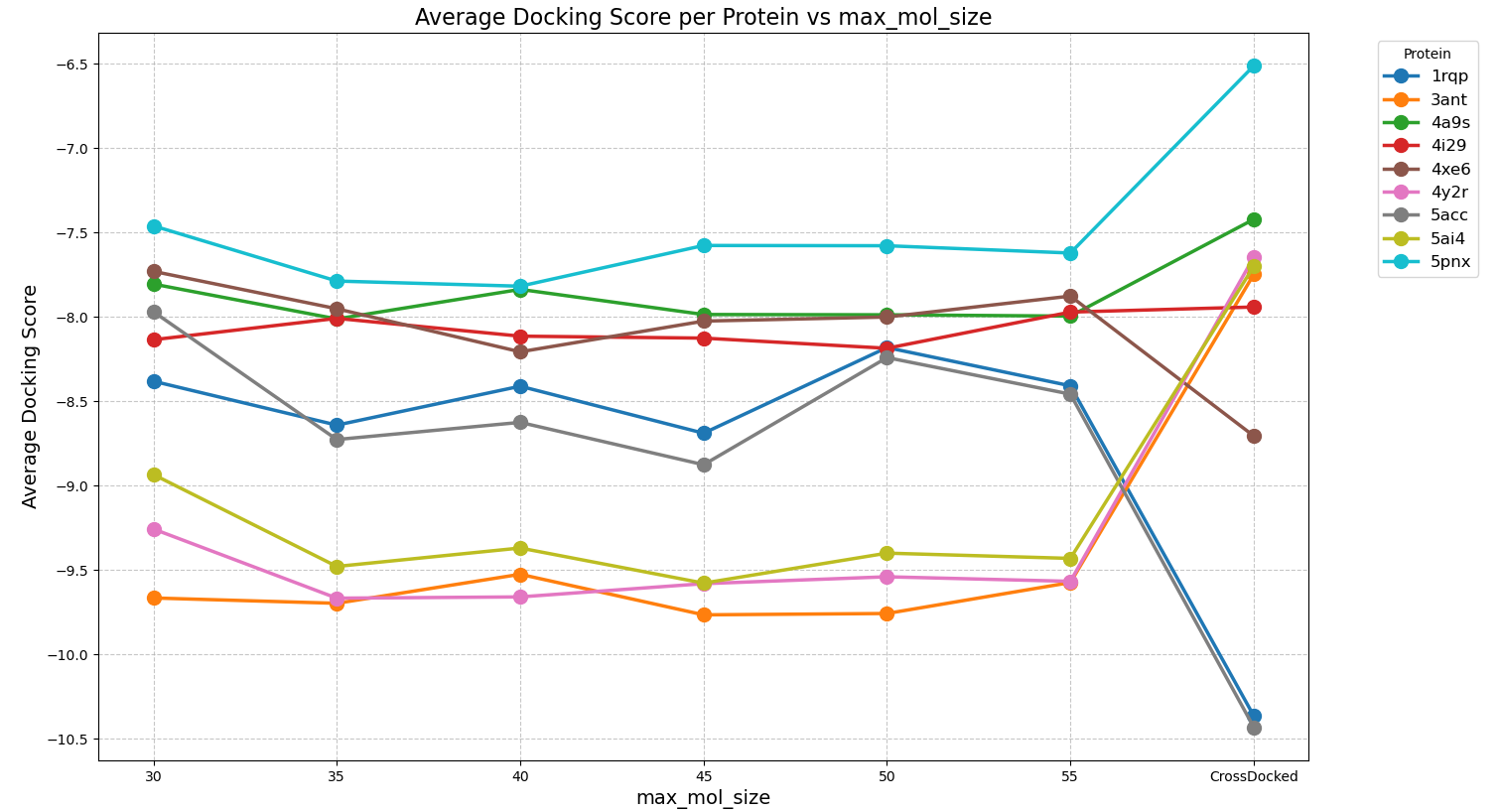}
    \includegraphics[width=1.0\linewidth]{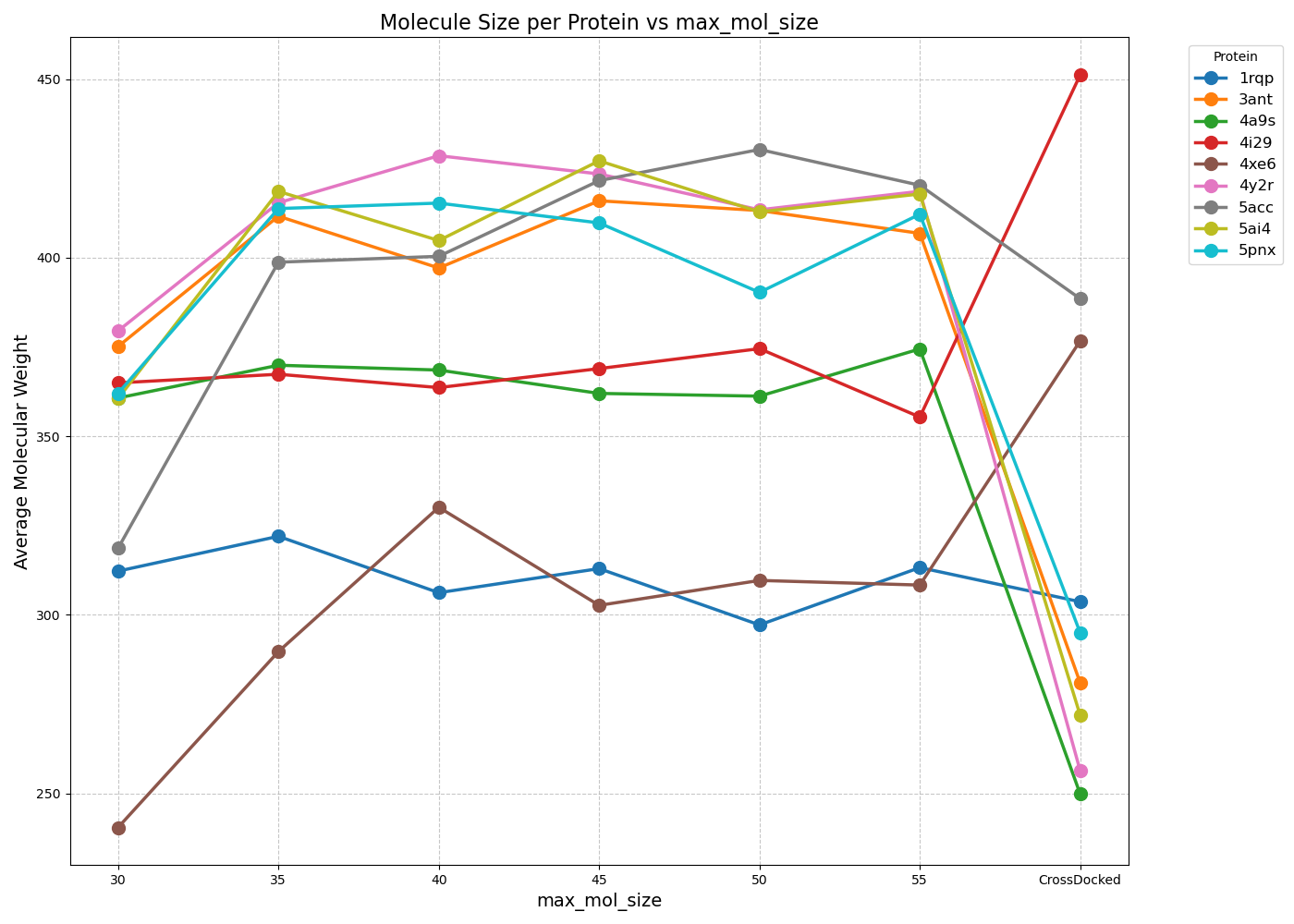}
    \caption{Average docking score and molecule weight under different $max\_mol\_size$}
    \label{fig:max_mol_size}
\end{figure}

After $max\_assay\_num$ is fixed to 10, we vary $N_{mol}$ across [1, 3, 5, 8, 12] and find $N_{mol}=8$ yields the best results, as shown in Figure \ref{fig:N_mol}. Finally, we fix $max\_assay\_num$ to 10 and $N_{mol}$ to 8 and examine $max\_mol\_size$'s influence on the generated molecules. We do not explicitly prompt the LLM to generate molecules of a specific size. The generated molecules' size and the docking score both correlate with $max\_mol\_size$, as shown in Figure \ref{fig:max_mol_size}. Therefore, we set $max\_mol\_size$ to different values as a way to approximately control the generated molecules' sizes.
Additional figures used to guide the hyperparameter tuning are available in our GitHub repository.

For LLM selection, we first select GPT 4o as the base model, a closed weights model. Then, we select DeepSeekV3, an open weights model, so that Assay2Mol does not rely solely on proprietary LLMs. Additionally, we incorporate Gemma-3-27B, a high-performing open weights model that can be locally deployed on a single GPU. For the motivating example, we use the ChatGPT 4o web version. For GPT 4o, we use the version "chatgpt-4o-latest" (as of May 2025) for Assay2Mol experiments and version "gpt-4o-2024-11-20" for LLM BioAssay relevance assessment (Table \ref{tab:relevance-review}), which was completed earlier. For DeepSeekV3, we use the version "DeepSeek-V3-0325" for Assay2Mol and "DeepSeek-V3-1226" for relevance assessment  (Table \ref{tab:relevance-review}). For Gemma-3-27B, we run the 4-bit bitsandbytes quantized version provided by unsloth\footnote{\url{https://huggingface.co/unsloth/gemma-3-27b-it-unsloth-bnb-4bit}} on a single RTX 5090 GPU.

\subsection{LLM molecule validity and price}
\label{sec:llm_validity}
Though DeepSeekV3 outperforms GPT 4o in Table \ref{tab:crossdock_result}, DeepSeekV3 generates far more invalid molecules than GPT 4o. We define validity as the percentage of unique generated SMILES that are parsable by RDKit. We list the validity and API price of different LLMs in Table \ref{tab:validity_cost}. 

\begin{table}[ht]
    \centering
    \resizebox{0.5\textwidth}{!}{
    \begin{tabular}{c|c|c|c}
    \toprule
        LLM& Validity & Input Price & Output Price \\
        \hline
        Gemma-3-27B & 84.32 & N/A & N/A \\
        DeepSeekV3 & 75.84 &  \$0.07/\$0.27 & \$1.1  \\
        GPT 4o & 94.33 & \$2.5/\$5 & \$20  \\
        \bottomrule
    \end{tabular}
    }
    \caption{Validity of generated molecules and API price of different LLMs. The price is for 1M input/output tokens. The numbers in input price represent the cost of generation with a cache hit and cache miss, respectively. The price for Gemma-3-27B is not applicable because it is run on a local machine.}
    \label{tab:validity_cost}
\end{table}

\begin{figure}
    \centering
    \includegraphics[width=1.0\linewidth]{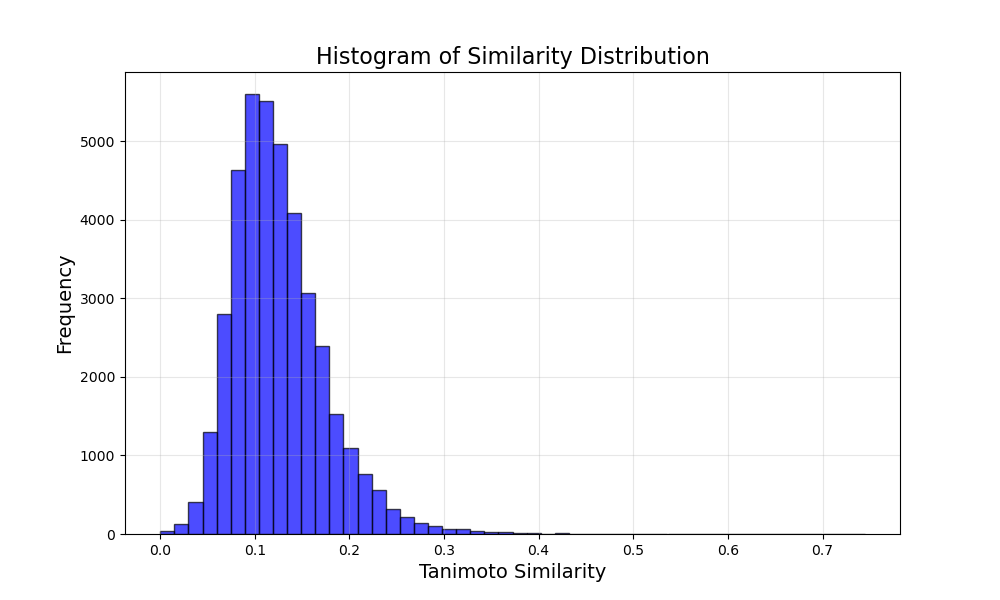}
     \caption{Similarity distribution between generated molecules and high docking score context molecules.}
    \label{fig:similarity_hist}
\end{figure}

\begin{figure}[ht]
    \centering
    \includegraphics[width=1.0\linewidth]{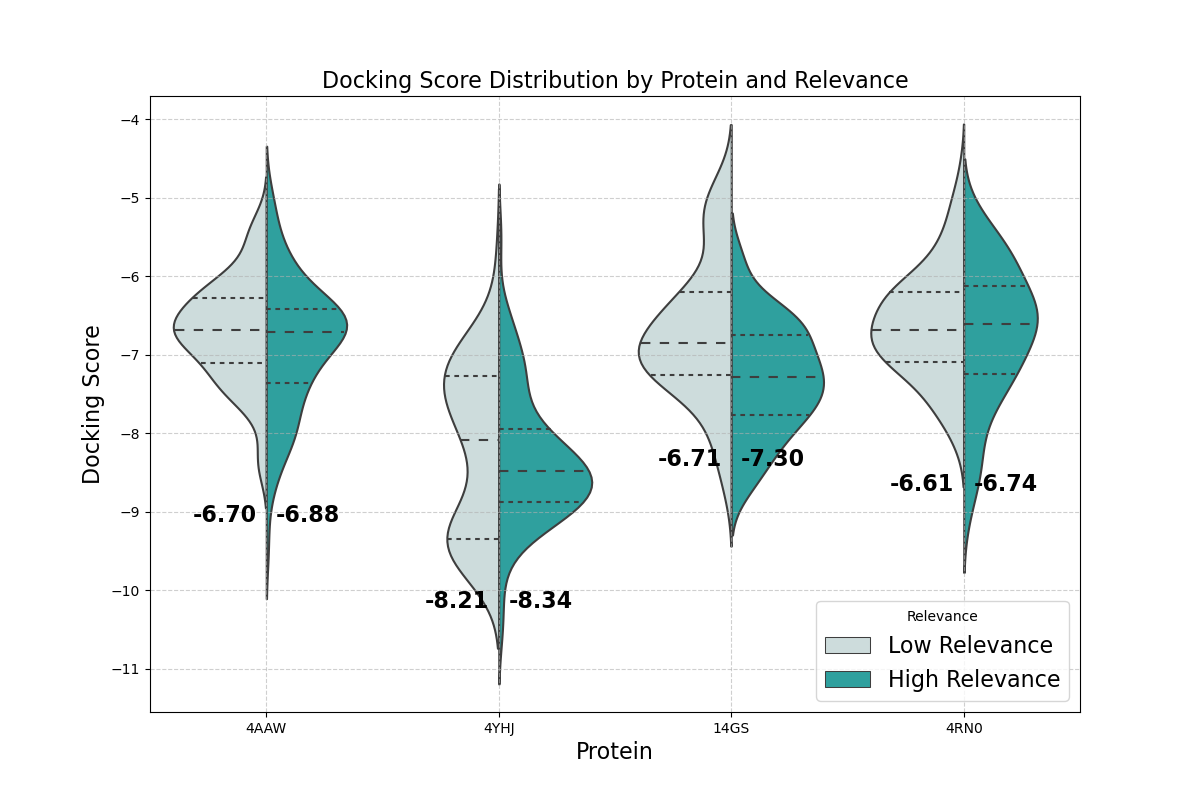}
    \caption{Distribution of docking scores across four proteins using high- and low-relevance BioAssays as context. The average docking score is shown on the top of each plot.}
    \label{fig:dissimilar_plot}
\end{figure}

\subsection{Similarity analysis}
\label{similarity_analysis}
Given the molecules provided in the BioAssay context, one natural question arises: Does the LLM simply copy and slightly modify the active molecules, rather than generating novel ones?
To address this question, we dock the context molecules against five proteins (4AAW, 4YHJ, 14GS, 2V3R, 4RN0) and analyze the similarity between the generated molecules and the context molecules. We select context molecules with docking scores more favorable than the reference molecules and consider them to be high-scoring context molecules. Then, we compute the Tanimoto similarity between all pairs of the high-scoring context molecules and the generated molecules using Morgan fingerprints as features (Figure \ref{fig:similarity_hist}). Most generated molecules are dissimilar to the high-scoring context molecules, suggesting that the LLM is learning from the context rather than simply making slight modifications to the context molecules.

\subsection{BioAssay context relevance experiment}
Table \ref{tab:normalized_table} already shows that Assay2Mol performs worse on proteins that do not have relevant BioAssays. We further investigate the impact of context relevance by substituting relevant BioAssays with irrelevant ones. Specifically, we select four proteins for which relevant BioAssays can be retrieved: PDB IDs 4AAW, 4YHJ, 14GS, and 4RN0. During the retrieval stage, we replace the BioAssays with those having relevance scores (measured in cosine similarity) ranging from 0.4 to 0.5, scores much lower than the relevant BioAssays, while keeping the rest of the Assay2Mol pipeline unchanged.
The result is shown in Figure \ref{fig:dissimilar_plot}. Although high-relevance BioAssays consistently outperform low-relevance BioAssays on average, the difference is not particularly pronounced. We suspect this is because GPT 4o also demonstrates the capability to generate molecules solely based on protein descriptions.

To investigate further, we analyzed the context molecules from the irrelevant BioAssays selected by GPT 4o as templates and examined their docking scores toward the query protein.
Across the targets 4AAW, 4RN0, 14GS, and 4YHJ the average value of the docking scores are -6.61, -6.04, -5.91, and -6.69, respectively.
Even when the BioAssays were irrelevant to the query protein, these context molecules can exhibit moderately high docking scores, making it less surprising that molecules generated from that context yield comparable docking scores.

\begin{figure*}
    \centering
    \includegraphics[width=1.0\linewidth]{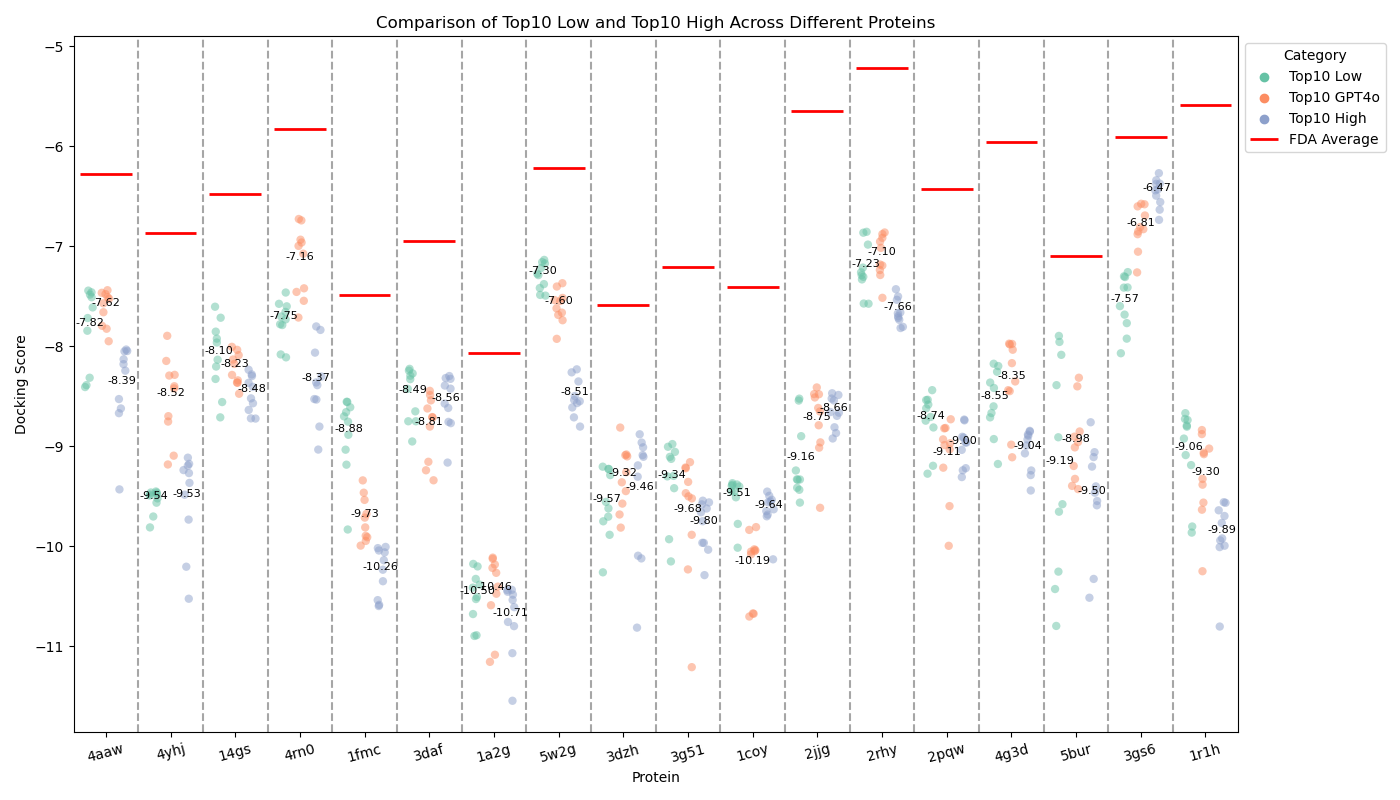}
    \centering
    \caption{Distribution of the top 10 docking scores from molecules with high- and low-relevance BioAssays as context for different proteins.}
    \label{fig:relevance_top}
\end{figure*}

However, we noticed that the lower tails of the violin plots representing high-relevance molecules tend to extend longer than those of the low-relevance group (Figure \ref{fig:dissimilar_plot}), indicating that the best docking scores are more frequently found among high-relevance molecules. When using docking for hit identification, researchers focus on the top-scoring molecules as opposed to the entire distribution because these are most likely to be enriched for actual active molecules. Thus, we extended the experiment to 18 proteins and only focused on the molecules with the 10 best docking scores per model. We found that in 11 out of 18 proteins, the high-relevance group performed better than both the low-relevance group and the without context group, which is the GPT 4o baseline (Figure \ref{fig:relevance_top}). In five proteins (4YHJ, 3DAF, 3DZH, 1COY, 2PQW), the high-relevance group performed similarly to other groups. In two proteins (2JJG, 3GS6), the high-relevance group performed worse than other groups.

These results suggest that in some cases, molecules derived from low-relevance BioAssays can still achieve favorable docking scores either by chance or due to biases in the docking software. Nonetheless, molecules from high-relevance BioAssays generally demonstrate superior performance when focusing on the tail of the docking score distribution.

\subsection{Antagonistic protein pair}
\label{TFPI discussion}
During the exploration of BioAssay retrieval, we found a scenario that is beyond the current capabilities of Assay2Mol. We tested Assay2Mol with recent, real drug targets by selecting the most recently FDA-approved drug of 2024\footnote{\url{https://www.fda.gov/drugs/novel-drug-approvals-fda/novel-drug-approvals-2024}}, concizumab-mtci.
This drug is an inhibitor of the human Tissue Factor Pathway Inhibitor (TFPI)\footnote{\url{https://pubchem.ncbi.nlm.nih.gov/protein/P10646}} protein. Using the protein description of TFPI from PubChem, we performed the BioAssay retrieval task. The top-matching BioAssay\footnote{\url{https://pubchem.ncbi.nlm.nih.gov/bioassay/1381622}} related to Tissue Factor (TF) and Coagulation Factor VII (FVII). Part of the retrieval result is shown in Table \ref{tab:TFPI_retrieval}. TFPI, TF, and FVII are key regulators of the extrinsic coagulation pathway, which is responsible for initiating blood clotting \citep{mcvey_tissue_1999}.
TF initiates coagulation, FVII promotes the process, and TFPI inhibits it, maintaining the dynamic balance of the coagulation system. TF inhibitors are designed to treat hypercoagulability and thrombosis. TFPI inhibitors are used for treating hemophilia A and B. Since TF and TFPI have antagonistic functions, an incorrect selection of BioAssays leads to misaligned molecule design.
When designing an inhibitor for TFPI, retrieving BioAssays targeting TF directs the molecule design in the opposite direction. Among the 300 retrieved BioAssays, only four directly relate to TFPI, and their target protein is mouse TFPI. The majority are associated with TF, the functional opposite of TFPI.

Although reasoning LLMs (for example, GPT o1 and DeepSeek R1) recognize the antagonistic relationship between the retrieved BioAssays and the query protein—allowing them to discard most mismatched results—the process is not completely accurate. Some of the 296 TF BioAssays are still incorrectly labeled as relevant to TFPI, representing false positives. The LLMs still learn from inhibitors of the opposing protein, leading to failure in molecule design. While the CrossDocked dataset contains no such cases, developing a more robust retrieval pipeline remains a priority for future work.

\begin{table*}[htbp]
\begin{tabular}{llp{1.5cm}p{1.7cm}p{1.7cm}p{2cm}p{2.3cm}}
\hline
\textbf{Relevance} & \textbf{Target} & \textbf{GPT 4o errors} & \textbf{GPT 4o accuracy (\%)} & \textbf{DeepSeek-V3 errors} & \textbf{DeepSeek-V3 accuracy (\%)} & \textbf{GPT 4o accuracy $-$ DeepSeek-V3 accuracy (\%)} \\
\hline
high & 5W2G & 1 & 90 & 1 & 90 & 0 \\
high & 3G51 & 4 & 60 & 10 & 0 & 60 \\
high & 1COY & 3 & 70 & 9 & 10 & 60 \\
high & 2JJG & 5 & 50 & 9 & 10 & 40 \\
high & 2RHY & 1 & 83 & 3 & 50 & 33 \\
high & 2PQW & 0 & 100 & 2 & 67 & 33 \\
high & 4G3D & 5 & 50 & 5 & 50 & 0 \\
medium & 4AAW & 4 & 60 & 6 & 40 & 20 \\
medium & 4YHJ & 0 & 100 & 1 & 90 & 10 \\
medium & 14GS & 1 & 90 & 9 & 10 & 80 \\
medium & 4RN0 & 1 & 90 & 4 & 60 & 30 \\
medium & 1FMC & 0 & 100 & 10 & 0 & 100 \\
medium & 3DAF & 0 & 100 & 10 & 0 & 100 \\
medium & 1A2G & 0 & 100 & 10 & 0 & 100 \\
medium & 3DZH & 0 & 100 & 8 & 20 & 80 \\
medium & 5BUR & 0 & 100 & 10 & 0 & 100 \\
low & 1R1H & 0 & 100 & 7 & 30 & 70 \\
low & 5B08 & 0 & 100 & 3 & 70 & 30 \\
low & 5I0B & 1 & 90 & 4 & 60 & 30 \\
low & 3KC1 & 0 & 100 & 4 & 60 & 40 \\
low & 1D7J & 1 & 90 & 4 & 60 & 30 \\
no & 2Z3H & 0 & 100 & 0 & 100 & 0 \\
no & 2V3R & 0 & 100 & 0 & 100 & 0 \\
no & 3B6H & 8 & 20 & 10 & 0 & 20 \\
no & 4P6P & 0 & 100 & 0 & 100 & 0 \\
\hline
\textbf{total} & 25 & 35 & 86 & 139 & 43 & 43
\end{tabular}
\caption{Manual review of LLM BioAssay relevance assessment by a single expert computational chemist (author S.S.E.). All targets have 10 BioAssays retrieved except for 2RHY and 2PQW, which have 6 each. The reason is there are only 6 BioAssays remained after filtering.}
\label{tab:relevance-review}
\end{table*}

\begin{table*}[htbp]
    \centering
    \begin{tabular}{cccc}
        \toprule
        \textbf{BioAssay} & \textbf{Score} & \textbf{Target} & \textbf{Relationship} \\
        \midrule
        \href{https://pubchem.ncbi.nlm.nih.gov/bioassay/1381622}{1381622}  & 0.8284  & Tissue factor, Coagulation factor VII & Opposite \\
        \href{https://pubchem.ncbi.nlm.nih.gov/bioassay/51307}{51307}    & 0.8245  & Tissue factor, Coagulation factor VII & Opposite \\
        \href{https://pubchem.ncbi.nlm.nih.gov/bioassay/397857}{397857}   & 0.8233  & Tissue factor, Coagulation factor VII & Opposite \\
        \href{https://pubchem.ncbi.nlm.nih.gov/bioassay/385437}{385437}   & 0.8202  & Tissue factor & Opposite \\
        \href{https://pubchem.ncbi.nlm.nih.gov/bioassay/385438}{385438}   & 0.8194  & Tissue factor & Opposite \\
        \href{https://pubchem.ncbi.nlm.nih.gov/bioassay/385435}{385435}   & 0.8190  & Tissue factor & Opposite \\
        \href{https://pubchem.ncbi.nlm.nih.gov/bioassay/385436}{385436}   & 0.8188  & Tissue factor & Opposite \\
        \href{https://pubchem.ncbi.nlm.nih.gov/bioassay/1871821}{1871821}  & 0.8183  & Coagulation factor VII & Opposite \\
        \href{https://pubchem.ncbi.nlm.nih.gov/bioassay/360094}{360094}   & 0.8182  & Tissue factor & Opposite \\
        \href{https://pubchem.ncbi.nlm.nih.gov/bioassay/385434}{385434}   & 0.8180  & Tissue factor & Opposite \\
        \midrule
        \multicolumn{4}{c}{$\vdots$} \\ 
        \midrule
        \href{https://pubchem.ncbi.nlm.nih.gov/bioassay/1620662}{1620662}  & 0.7941  & Tissue factor pathway inhibitor & Correct \\
        \href{https://pubchem.ncbi.nlm.nih.gov/bioassay/1620663}{1620663}  & 0.7903  & Tissue factor pathway inhibitor & Correct \\
        \midrule
        \multicolumn{4}{c}{$\vdots$} \\ 
        \midrule
        \href{https://pubchem.ncbi.nlm.nih.gov/bioassay/1476906}{1476906}  & 0.7888  & Coagulation factor XII & Opposite \\
        \href{https://pubchem.ncbi.nlm.nih.gov/bioassay/1775843}{1775843}  & 0.7888  & Carboxypeptidase B2 & Opposite \\
        \href{https://pubchem.ncbi.nlm.nih.gov/bioassay/304051}{304051}   & 0.7888  & Carboxypeptidase B2 & Opposite \\
        \href{https://pubchem.ncbi.nlm.nih.gov/bioassay/52023}{52023}    & 0.7888  & Coagulation factor X & Opposite \\
        \href{https://pubchem.ncbi.nlm.nih.gov/bioassay/1426475}{1426475}  & 0.7888  & Coagulation factor X & Opposite \\
        \href{https://pubchem.ncbi.nlm.nih.gov/bioassay/1775842}{1775842}  & 0.7888  & Carboxypeptidase B2 & Opposite \\
        \href{https://pubchem.ncbi.nlm.nih.gov/bioassay/212091}{212091}   & 0.7887  & Tissue factor & Opposite \\
        \href{https://pubchem.ncbi.nlm.nih.gov/bioassay/362212}{362212}   & 0.7887  & TISSUE: Plasma & Opposite \\
        \href{https://pubchem.ncbi.nlm.nih.gov/bioassay/1382994}{1382994}  & 0.7886  & Coagulation factor X & Opposite \\
        \href{https://pubchem.ncbi.nlm.nih.gov/bioassay/1657762}{1657762}  & 0.7886  & TISSUE: Plasma & Opposite \\
        \bottomrule
    \end{tabular}
    \caption{BioAssay retrieval result using the TFPI description as input. The Score is measured with cosine similarity. Only four out of 300 BioAssays are related to TFPI while others have the opposite function in the pathway.}
    \label{tab:TFPI_retrieval}
\end{table*}

\subsection{Software}
We use the hERG Blocker classifier from the ADMETlab 3.0 web server \citep{10.1093/nar/gkae236} to predict the hERG score, which is in the range [0,1].\\
We use AutoDock Vina \citep{doi:10.1021/acs.jcim.1c00203} v1.5.7 as the docking software.  We use the original ligand center from the protein-ligand complex as the docking center, set the box size to 20 Å in each dimension (x, y, z), and set the exhaustiveness to 32.

\subsection{Assay summarization example}
\label{summarization example}
\begin{tcolorbox}[enhanced, breakable, colback=gray!5, colframe=gray!80, width=\columnwidth, title=Summarized BioAssay Example]
This BioAssay measures the inhibition of GRK5-mediated phosphorylation of rhodopsin in bovine rod outer segment membranes under white light conditions. It evaluates the efficacy of cyclic peptide inhibitors derived from the HJ loop of GRK2, providing insights into their potency and selectivity.

This is the activity data table. Each line has the SMILES, followed by activity type (active, inactive or unspecified) and the experimental value.

\texttt{[C@H](C(=O)N1)CCCCN))...(N)N)CC(C)C} \textcolor{blue}{Unspecified} \quad Inhibition <5\%

\texttt{CC[C@H](C)[C@@H](C(=O)O)...NC(=O)CN} \textcolor{blue}{Unspecified} \quad Inhibition <5\%

\vspace{0.5em}

This BioAssay evaluates the inhibition of recombinant human GRK5 expressed in Sf9 insect cells, measuring the decrease in phosphorylation of urea-washed bovine rod outer segments in the presence of Gbetagamma subunits and [gamma-32P]-ATP.

This is the activity data table. Each line has the SMILES, followed by activity type (active, inactive or unspecified) and the experimental value.

\texttt{CC[C@H](C)[C@@H](C(=O)N...(CCCCN)N} \textcolor{green}{Active} \quad IC50 = 2100 nM

\texttt{\detokenize{C1=CC=C(C=C1)/C=C\2/C3=CC=CC=C3C(=O)N2}} \textcolor{red}{Inactive} \quad IC50 = 60000 nM
\end{tcolorbox}

\subsection{Prompts}

\subsubsection{BioAssay summarization prompt}
\label{BioAssay summarization Prompt}
\begin{tcolorbox}[breakable, colback=gray!5, colframe=gray!80, width=\columnwidth, title=Prompt for BioAssays Summarization]

\paragraph{\textbf{Instruction:}}
You are an expert in \textbf{BioAssay analysis} and \textbf{data extraction}. Your task is to carefully analyze the provided BioAssay JSON data and extract structured key information, including:

1. \textbf{BioAssay Summarization} – A concise summary of what this assay measures and its scientific purpose.\\
2. \textbf{Assay Type} – The experimental technique used (e.g., \textbf{Enzymatic Inhibition, Fluorescence Assay, SPR, Radioligand Binding}).\\
3. \textbf{Summary of Observations} – Important scientific insights derived from the BioAssay, including key patterns in activity, structural features affecting activity, and notable findings.

\paragraph{\textbf{Step-by-step extraction process:}}\mbox{}\\
- Parse the \textbf{"descr"} section of the JSON, identifying key information about the assay. If this BioAssay is a counterscreen assay, set "CounterScreen" to "True". \\
- Identify the \textbf{Assay Type} by analyzing the \textbf{"name"} field.\\
- Extract \textbf{scientific insights} from the description and comments to create the \textbf{Summary of Observations}.\\
- Generate a \textbf{concise and informative summary} of the BioAssay, keeping scientific accuracy and relevance.

\paragraph{\textbf{Output Format}}
Return the extracted data in the following structured format:\\

\begin{lstlisting}[basicstyle=\ttfamily\small, breaklines=true]
json
{
  "BioAssay_Summary": "A brief but complete summary of what this assay is measuring and why it is important.",
  "Assay_Type": "The experimental method used (e.g., Enzymatic Inhibition, Fluorescence, SPR, etc.)",
  "Summary_of_Observations": "Scientific insights, key findings, and notable trends from the BioAssay.",
  "CounterScreen": "True" if the BioAssay is identified as CounterScreen against Query Protein, else "False".
}
\end{lstlisting}
\paragraph{\textbf{Query Protein:}}\mbox{}\\
\{Protein Description\}
\paragraph{\textbf{BioAssay JSON}}\mbox{}\\
\{BioAssay JSON\}
\end{tcolorbox}

\subsubsection{Generation prompt}
\label{Generation Prompt}
\begin{tcolorbox}[breakable, colback=gray!5, colframe=gray!80, width=\columnwidth, title=Molecule Generation Prompt]
\paragraph{Role: AI Molecular Generator and BioAssay Analyst} \mbox{} \\

\paragraph{Profile} \mbox{}\\
- \textbf{Author}: LangGPT \\
- \textbf{Version}: 1.1 \\ 
- \textbf{Language}: English \\
- \textbf{Description}: An AI model specialized in analyzing BioAssay results, understanding protein-ligand interactions, and generating high-affinity molecules based on experimental data. \\

\paragraph{Skills} \mbox{}\\
- Understanding \textbf{protein-ligand interactions} from experimental BioAssay data. \\
- Interpreting \textbf{BioAssay results} and extracting meaningful insights. \\
- Learning from \textbf{high-affinity molecules} in BioAssay data to generate new molecules. \\
- Ensuring \textbf{high binding affinity and specificity}, while avoiding \textbf{Pan-assay interference compounds (PAINS)}. \\
- Generating drug-like molecules that align with known \textbf{active reference compounds}. \\

\paragraph{Rules} \mbox{}\\
1. \textbf{Carefully analyze} the input description of the \textbf{protein}, \textbf{BioAssay}, and \textbf{experimental results}. \\
2. Identify \textbf{high-affinity molecules} (low IC50/Kd values) from the \textbf{BioAssay data} as \textbf{reference molecules}. \\
3. Use reference molecules to \textbf{learn key functional groups and molecular scaffolds}. \\
4. Focus on \textbf{specificity rather than only high docking scores}. \\
5. \textbf{Each generated molecule should be enclosed within [BOS] and [EOS]}. \\
6. \textbf{Each SMILES should be numbered from 1 to 10, with one per line.} \\
7. Avoid \textbf{PAINS compounds} and prioritize \textbf{drug-likeness}. \\
8. \textbf{Do not blindly maximize molecular size}, as larger molecules may have artificially high docking scores but poor specificity.

\paragraph{Workflows}

\paragraph{\textbf{Step 1: Understand the BioAssay and Its Relation to the Query Protein}}\mbox{}\\
- The BioAssays may or may not be related to the \textbf{Query Protein}, please identify the correct Query Protein first.\\
  - The type of \textbf{assay method} used (e.g., enzymatic, fluorescence, cell-based).
  - How the \textbf{assay measures protein-ligand interaction}.
  - The \textbf{affinity measurements} (e.g., IC50, Kd, Ki).\\
- Extract \textbf{key active molecules} from BioAssay results. (e.g. IC50, Ki, Kd<100nM)
- Identify molecular features that contribute to \textbf{high binding affinity}. \\

\paragraph{\textbf{Step 2: Learn from Active Molecules and Think Step by Step}} \mbox{} \\
- Extract \textbf{key functional groups} and \textbf{molecular scaffolds} from high-affinity reference molecules. \\
- Avoid \textbf{PAINS compounds} and prioritize \textbf{specificity}. \\
- Ensure molecules remain within a \textbf{reasonable drug-like chemical space}. \\
- Optimize molecular properties for \textbf{binding affinity and selectivity}. \\

\paragraph{\textbf{Step 3: Generate 10 High-Affinity Molecules}} \mbox{}\\
- Use the \textbf{active reference molecules} as a learning guide, and use the \textbf{low binding affinity molecules} as negative samples. \\
- Each generated molecule should be optimized for \textbf{binding affinity and specificity}.
- The output format must follow this structure:\\
  - Each \textbf{SMILES string should be enclosed in [BOS] and [EOS]}. \\
  - Each SMILES should be \textbf{numbered from 1 to 10}, with each on a separate line.\\
  - Avoid \textbf{PAINS compounds} and prioritize \textbf{drug-likeness}.\\
  - Avoid generating molecules that are too large

\paragraph{\textbf{Step 4: Justify the Molecular Selection}}
- Explain how the \textbf{reference molecules} influenced the molecular design. \\
- Describe how the \textbf{assay results} guided molecular modifications. \\
- Justify why these molecules should have \textbf{high binding affinity and specificity}. \\

---

\paragraph{Output Format:}

\paragraph{\textbf{1. BioAssay Understanding \& Analysis}} \mbox{} \\
- Step-by-step reasoning about the BioAssay, its setup, and its relevance to the query protein \\
\paragraph{\textbf{2. Selected Reference Molecules from BioAssay}} \mbox{} \\
- List of highly active molecules from BioAssay used as reference. \\
\paragraph{\textbf{3. Generated Molecules}} \mbox{}\\
{\texttt{[BOS] SMILES\_1 [EOS]}}\\
$\vdots$\\
{\texttt{[BOS] SMILES\_10 [EOS]}}\\
\paragraph{\textbf{4. Justification for Molecular Selection}} \mbox{}\\
- Explanation of how reference molecules influenced design choices, ensuring specificity and affinity while avoiding PAINS.\\

\paragraph{\textbf{Query Protein}} \mbox{}\\
\{Protein Description\} \\
\paragraph{\textbf{BioAssays}} \mbox{}\\
\{Assay Content\} \\
\end{tcolorbox}

\subsubsection{Relevance assessment prompt}
\label{Relevance Assessment Prompt}
\begin{tcolorbox}[breakable, colback=gray!5, colframe=gray!80, width=\columnwidth, title=Relevance Assessment Prompt]
\paragraph{\textbf{Instruction:}} \mbox{}\\
You are an expert in \textbf{BioAssay analysis} and \textbf{data extraction}. Your task is to carefully analyze the provided BioAssay JSON data and extract structured key information, and decide whether the protein studied in the BioAssay is related to the Query Protein, with broader consideration of similarity:\\
- If the BioAssay can help protein inhibitor design toward the \textbf{Query Protein}, set "Relevant": "True".\\
- \textbf{Use a broad definition of similarity}, including:\\
- Same \textbf{protein family} (e.g., GRK4 and GRK2 are both \textbf{G protein-coupled receptor kinases}).\\
- Structural similarity (e.g., homologous domains, catalytic site conservation).\\
- Functional similarity (e.g., overlapping substrates).

\paragraph{\textbf{Output Format}}\mbox{}\\
\begin{lstlisting}[basicstyle=\ttfamily\small, breaklines=true]
json
{
  "Relevant": "True"  // If the target in this BioAssay shares protein family, structure, function, or pathway with the {Query Protein, else set to "False". Only set to "False" if there is no meaningful similarity.
}
\end{lstlisting}

\paragraph{\textbf{Query Protein}}\mbox{}\\
\{protein description\}
\paragraph{\textbf{BioAssay JSON}}\mbox{}\\
\{BioAssay content\}
\end{tcolorbox}

\subsubsection{Ablation study generation prompt}
\begin{tcolorbox}[breakable, colback=gray!5, colframe=gray!80, width=\columnwidth, title=Ablation study prompt]
I would like to design drug-like small molecules with high binding affinity to a specific protein. \texttt{\{protein\_description\}}  Think step by step, generate 10 unique SMILES strings, try to generate diverse molecules instead of making slight change on current molecule. Each molecule should have `[BOS]' at the beginning and `[EOS]' at the end. Each SMILES should be numbered from 1 to 10 and on a separate line.
\label{ablation_prompt}
\end{tcolorbox}

\subsubsection{Molecule optimization prompt}
\label{hERG prompt}
\begin{tcolorbox}[breakable, colback=gray!5, colframe=gray!80, width=\columnwidth, title=Molecule optimization prompt]
To enhance molecular specificity and minimize off-target effects, we aim to reduce potential activity against the hERG channel.
\{hERG description\}\\
\{hERG BioAssays\}\\
Given the retrieved BioAssays for the hERG channel and the associated activity data table, identify molecular features commonly associated with low activity as favorable and those associated with high activity as undesirable. Using this information, optimize the following ten candidate SMILES strings to reduce their likelihood of interacting with the target.
\{Input SMILES\}\\
The output should follow the same format: ten optimized SMILES strings, each enclosed in [BOS] and [EOS], with numbering from 1 to 10.
\end{tcolorbox}
\end{document}